%% file: colm2024_conference.tex
\documentclass[dvipsnames]{article}
\usepackage{colm2024_conference}

\usepackage{booktabs}
\usepackage{enumitem}
\usepackage{wrapfig}
\usepackage{algorithm}
\usepackage{algpseudocode}
\usepackage{graphicx}
\usepackage[misc]{ifsym}
\usepackage{pifont}
\usepackage{needspace}

\usepackage{microtype}
\usepackage{amsmath}
\usepackage{colortbl}
\usepackage[utf8]{inputenc}
\usepackage[T1]{fontenc}
\definecolor{lightgray}{rgb}{0.9,0.9,0.9}
\usepackage{caption}
\usepackage{subcaption}
\usepackage{setspace}
\usepackage{tcolorbox}
\tcbuselibrary{skins,listings,breakable}
\usepackage{url}
\usepackage{multirow}
\usepackage{tcolorbox}
\usepackage{tabularx}
\usepackage{blindtext}
\usepackage{pgfplots}
\pgfplotsset{compat=1.18}
\usepackage{tikz}
\usetikzlibrary{er,positioning,bayesnet}
\usepackage{makecell}
\usepackage{tipa}
\usepackage{siunitx}
\usepackage{nicefrac}
\usepackage{listings}
\usepackage{xltabular}
\usepackage{adjustbox}
\usepackage{xurl}
\usepackage{rotating}
\usepackage[normalem]{ulem}

\usepackage{amssymb}
\usepackage{mathtools}
\usepackage{amsthm}
\usepackage{multicol}
\usepackage{inconsolata}
\usepackage[table]{xcolor}
\usepackage{cuted}
\usepackage{capt-of}

\definecolor{myyellow}{rgb}{.99,.94,.82}

\usepackage{booktabs}
\usepackage{multirow}
\usepackage[table]{xcolor}
\usepackage{makecell}
\usepackage{subcaption}

\definecolor{darkgreen}{RGB}{0,140,0}
\definecolor{rowgray}{RGB}{245,245,245}

\definecolor{myboxcolor}{RGB}{245,245,245} 
\definecolor{myframe}{RGB}{0,0,128} 

\newtcolorbox{mybody}{
  colback=myboxcolor,
  colframe=myframe,
  boxrule=1pt, % Adjust the border thickness
  left=1pt,
  right=1pt,
  top=1pt,
  bottom=1pt,
}

\useunder{\uline}{\ul}{}

\input{math_commands.tex}

\newcommand*\justify{%
  \fontdimen2\font=0.4em% interword space
  \fontdimen3\font=0.2em% interword stretch
  \fontdimen4\font=0.1em% interword shrink
  \fontdimen7\font=0.1em% extra space
  \hyphenchar\font=`\-% allowing hyphenation
}

\renewcommand{\texttt}[1]{%
  \begingroup
  \ttfamily
  \begingroup\lccode`~=`/\lowercase{\endgroup\def~}{/\discretionary{}{}{}}%
  \begingroup\lccode`~=`[\lowercase{\endgroup\def~}{[\discretionary{}{}{}}%
  \begingroup\lccode`~=`.\lowercase{\endgroup\def~}{.\discretionary{}{}{}}%
  \catcode`/=\active\catcode`[=\active\catcode`.=\active
  \justify\scantokens{#1\noexpand}%
  \endgroup
}

\usepackage{makecell}
\usetikzlibrary{tikzmark}
\makeatletter
\newcommand*\myfontsize{%
  \@setfontsize\myfontsize{7}{8}%
}
\makeatother

\definecolor{uclablue}{RGB}{159, 195, 224}

\definecolor{uclagold}{RGB}{255, 240, 180}

\definecolor{aliceblue}{RGB}{255, 238, 241}

\definecolor{cadmiumgreen}{rgb}{0.0, 0.42, 0.24}

\definecolor{myred}{rgb}{0.7, 0.3, 0.0}
\definecolor{myblue}{rgb}{0.2, 0.3, 0.6}
\definecolor{babygreen}{rgb}{0.85, 0.97, 0.85}

\definecolor{purple1}{RGB}{0, 0, 0}
\definecolor{purple2}{RGB}{0, 0, 0}
\definecolor{purple3}{RGB}{0, 0, 0}
\definecolor{purple4}{RGB}{0, 0, 0}

\definecolor{deepblue}{RGB}{48, 58, 82}

\definecolor{deepPurple}{HTML}{000000}
\definecolor{uclablue_old}{rgb}{0, 0, 0}
\hypersetup{
    breaklinks,
    citecolor=citegray,
    colorlinks=true,
    linkcolor=citegray,
    urlcolor=citegray
}

\newcommand{\deltaCell}[1]{%
  \ifdim #1 pt > 0pt
    \textcolor{darkgreen}{#1}%
  \else
    \ifdim #1 pt < 0pt
      \textcolor{red}{#1}%
    \else
      #1%
    \fi
  \fi
}
\newtcolorbox{mybox}[2][]
  {colback = black!5!white, colframe = black!75!black, fonttitle = \bfseries,
    colbacktitle = black!100!black, enhanced, before upper={\fontsize{8}{11}\obeyspaces\obeylines\selectfont}, fontupper=\selectfont,
    attach boxed title to top left={yshift=-2.2mm,xshift=4mm},
    title=#2,#1}

\title{SaaS-Bench: Can Computer-Use Agents Leverage Real-World SaaS to Solve Professional Workflows?}

\author{
Kean Shi$^{1,2,*}$~, Zihang Li$^{2,*}$, Tianyi Ma$^{1,*}$, Zengji Tu$^{1,2}$, Jialong Wu$^{1,2}$, Wendong Xu$^{1}$, Xinbo Xu$^{1,2}$, Qingyao Yang$^{1,3}$, Ruoyu Wu$^{1,2}$, Weichu Xie$^{2}$, Ming Wu$^{4}$, Jason Zeng$^{4}$, Michael~Heinrich$^{4}$, Elvis Zhang$^{5}$, Liang Chen$^{1,\dagger}$, Kuan Li$^{1,\dagger}$, Baobao Chang$^{2,\dagger}$
  \\[0.8em]
  {\fontsize{10pt}{11pt}\selectfont          % \large 放大字号；\bfseries 视需要加粗
  $^1$UniPat AI, $^2$PKU, $^3$HKU, $^4$0G Labs, $^5$Pipeline Lab%
}}

\definecolor{gold}{HTML}{EEB862}
\definecolor{forest}{HTML}{194E41}
\definecolor{saasorange}{HTML}{E88C03}
\definecolor{crimson}{HTML}{A2012A}

\newcommand{\cmark}{\textcolor{green!60!black}{\ding{51}}}
\newcommand{\xmark}{\textcolor{red!70!black}{\ding{55}}}
\newcommand{\pmark}{\textcolor{black}{$\triangle$}}

\newtcblisting{taskrawbox}[1]{
    enhanced,
    breakable,
    width=\linewidth,
    enlarge left by=0pt,
    enlarge right by=0pt,
    colback=gray!3,
    colframe=black!55,
    boxrule=0.5pt,
    arc=1mm,
    left=5pt,
    right=5pt,
    top=4pt,
    bottom=5pt,
    boxsep=0pt,
    title={#1},
    coltitle=red,
    fonttitle=\bfseries,
    colbacktitle=gray!8,
    toptitle=2pt,
    bottomtitle=2pt,
    lefttitle=5pt,
    righttitle=5pt,
    listing only,
    listing engine=listings,
    listing options={
        basicstyle=\ttfamily\tiny,
        breaklines=true,
        breakatwhitespace=false,
        breakautoindent=false,
        breakindent=0pt,
        postbreak={},
        columns=fullflexible,
        keepspaces=true,
        showstringspaces=false,
        xleftmargin=0pt,
        xrightmargin=0pt,
        resetmargins=true,
        aboveskip=0pt,
        belowskip=0pt,
        frame=none,
        literate={§}{{$\S$}}1
                 {—}{{-}}1
                 {"}{{"}}1
                 {"}{{"}}1
                 {'}{{'}}1
                 {'}{{'}}1
                 {…}{{\dots}}1
    }
}

\lstdefinestyle{taskrawstyle}{
    basicstyle=\ttfamily\tiny,
    breaklines=true,
    breakatwhitespace=false,
    breakautoindent=false,
    breakindent=0pt,
    postbreak={},
    columns=fullflexible,
    keepspaces=true,
    showstringspaces=false,
    xleftmargin=0pt,
    xrightmargin=0pt,
    resetmargins=true,
    aboveskip=0pt,
    belowskip=0pt,
    frame=none,
    literate={§}{{$\S$}}1
}

\newtcolorbox{taskassetbox}[1]{
    enhanced,
    breakable,
    width=\linewidth,
    colback=gray!3,
    colframe=black!55,
    boxrule=0.5pt,
    arc=1mm,
    left=5pt,
    right=5pt,
    top=4pt,
    bottom=5pt,
    boxsep=0pt,
    title={#1},
    coltitle=red,
    fonttitle=\bfseries,
    colbacktitle=gray!8,
    toptitle=2pt,
    bottomtitle=2pt,
    lefttitle=5pt,
    righttitle=5pt
}

\definecolor{trajbg}{RGB}{235,243,254}
\definecolor{trajframe}{RGB}{66,133,244}
\definecolor{verifybg}{RGB}{255,249,230}
\definecolor{verifyframe}{RGB}{230,150,0}

\newtcolorbox{trajbox}[1]{%
    enhanced, breakable,
    colback=trajbg, colframe=trajframe,
    boxrule=0.5pt, arc=1.5mm,
    left=6pt, right=6pt, top=5pt, bottom=5pt,
    title={\small #1},
    fonttitle=\bfseries,
    coltitle=white,
    colbacktitle=trajframe,
    toptitle=2pt, bottomtitle=2pt,
    lefttitle=6pt, righttitle=6pt,
}

\newtcolorbox{verifybox}[1]{%
    enhanced, breakable,
    colback=verifybg, colframe=verifyframe,
    boxrule=0.5pt, arc=1.5mm,
    left=6pt, right=6pt, top=5pt, bottom=5pt,
    title={\small #1},
    fonttitle=\bfseries,
    coltitle=white,
    colbacktitle=verifyframe,
    toptitle=2pt, bottomtitle=2pt,
    lefttitle=6pt, righttitle=6pt,
}

\begin{document}

\maketitle

\renewcommand{\thefootnote}{\fnsymbol{footnote}}
\footnotetext{$^*$Equal Core Contributors}
\footnotetext{$^\dagger$Correspondence: Liang~Chen~<\texttt{liangchen@unipat.ai}>, Kuan~Li~<\texttt{kuanli@unipat.ai}>, Baobao~Chang~<\texttt{chbb@pku.edu.cn}>}
\renewcommand{\thefootnote}{\arabic{footnote}}

% \begingroup
%   \renewcommand\thefootnote{\Letter}  % 把当前脚注编号改成 \Letter
%   \footnotetext{Correspondence: Liang Chen <liangchen@unipat.ai>, Kuan Li <kuanli@unipat.ai>} % 这一行会出现在本页最底部
% \endgroup

\newcommand{\datasetname}{\textsc{SaaS-Bench}\xspace}

\begin{abstract}
Computer-Using Agents (CUAs) are rapidly extending large language models (LLMs) beyond text-based reasoning toward action execution in more complex environments, such as web browsers and graphical user interfaces (GUIs). However, existing web and GUI agent benchmarks often rely on simplified settings, isolated tasks, or short-horizon interactions, making it difficult to assess capabilities of agents in realistic professional workflows. Software-as-a-Service (SaaS) environments are a natural choice for CUA evaluation, as they host a large share of modern digital work and naturally involve dynamic system states, cross-application coordination, domain-specific knowledge, and long-horizon dependencies. To this end, we introduce \textbf{\datasetname}, a benchmark built on 23 deployable SaaS systems across six professional domains, containing 106 tasks grounded in realistic work scenarios. These tasks require long-horizon execution, cover both text-only and multimodal settings, and are evaluated with weighted verification checkpoints that measure strict task completion and partial progress. Experiments show that representative LLM-based agents struggle on \datasetname, with even the strongest model completing fewer than 4\% of tasks end-to-end, exposing limitations in planning, state tracking, cross-application context maintenance, and error recovery. Code are available at \href{https://github.com/UniPat-AI/SaaS-Bench}{\github~UniPat-AI/SaaS-Bench} for reproduction.
\end{abstract}

% \begin{figure}[h]
%     \centering
%     \includegraphics[width=1.1\linewidth]{figures/abs_fig.pdf}
%     \caption{Caption}
%     \label{fig:abs_fig}
% \end{figure}

\begin{figure}[h]
    \centering
    \includegraphics[width=0.9\linewidth]{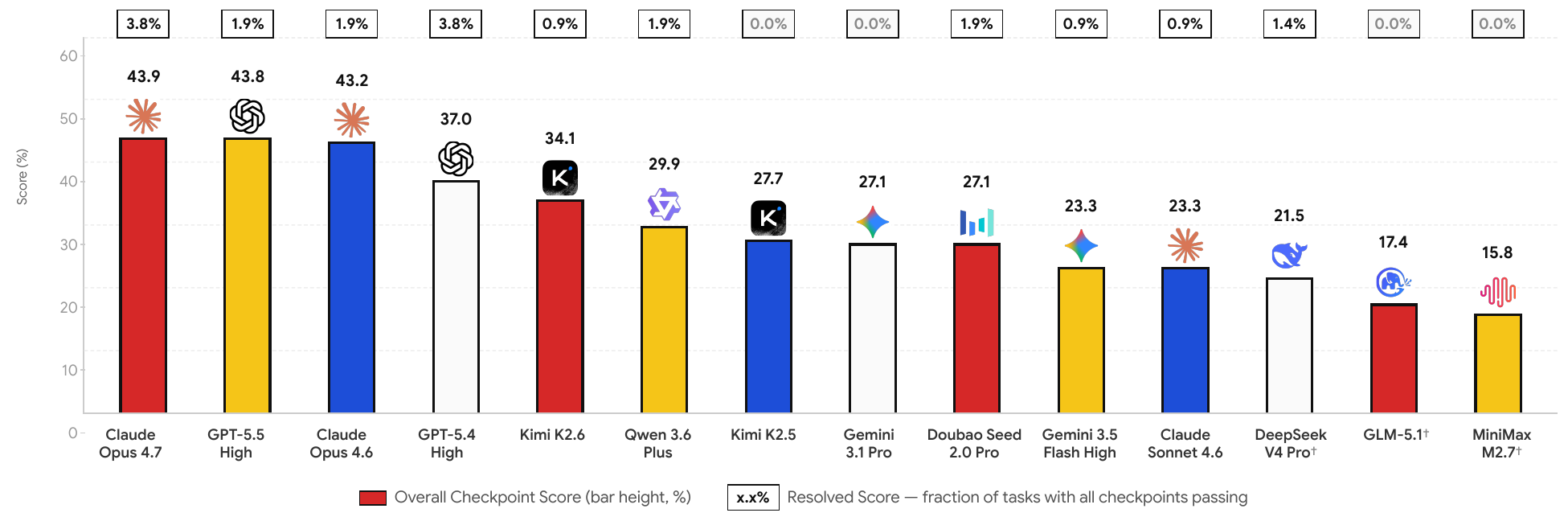}
    \caption{Leaderboard of \datasetname. We report overall checkpoint scores (bar height) and resolved scores for 14 frontier models across 106 long-horizon SaaS tasks. $^\dagger$donate text-only models.}
    \label{fig:abs_fig}
\end{figure}

\newpage

\input{contents/introduction}
\input{contents/related_work}
\input{contents/babyvision}
\input{contents/experiment}
\input{contents/discussion}
\input{contents/conclusion}

\clearpage
\bibliography{ref}
\bibliographystyle{colm2024_conference}

\clearpage
\input{contents/appendix}

\end{document}

%% file: math_commands.tex
\usepackage{amsmath,amsfonts,bm}

\def\eqref#1{equation~\ref{#1}}
\def\1{\bm{1}}

\DeclareMathAlphabet{\mathsfit}{\encodingdefault}{\sfdefault}{m}{sl}
\SetMathAlphabet{\mathsfit}{bold}{\encodingdefault}{\sfdefault}{bx}{n}

%% file: contents/introduction.tex
\section{Introduction}
\label{intro}

Recent advances in Large Language Models~(LLMs) have enabled the emergence of Computer-Using Agents (CUAs)~\cite{uitars,opencua,cua,computeruse}, which represent a paradigm shift from passive understanding to active execution~\cite{webarena,osworld,webvoyager}. Unlike traditional models that focus solely on language comprehension and generation, CUAs are capable of interacting with real-world software systems by taking actions across diverse interfaces, including graphical user interfaces, web browsers, and APIs. This enables them to accomplish end-to-end workflows, such as information retrieval, data manipulation, and multi-step task execution. As a result, evaluating the real capabilities of CUAs has become a central problem.

However, existing benchmarks fail to accurately reflect real-world agent capabilities, leading to systematic overestimation of performance.~\cite{mind2web,webarena,visualwebarena,osworld} First, they provide limited application-level complexity. Even when benchmarks use executable or self-hosted web environments, their page logic, backend constraints, and state transitions are often simpler than those of real SaaS systems. Second, they fail to capture real-world professional workflows, since tasks are typically confined to isolated, single-application scenarios with simplified goals, whereas real professional work naturally involves coordination across multiple systems, domain-specific knowledge, and structured multi-step processes. Third, they lack the type of long-horizon dependencies required by realistic SaaS workflows, where completing a task may involve over 100 interaction steps. Consequently, existing benchmarks provide limited insight into whether CUAs can effectively operate in realistic, high-value scenarios.

\begin{figure}
  \centering
  \includegraphics[width=\linewidth]{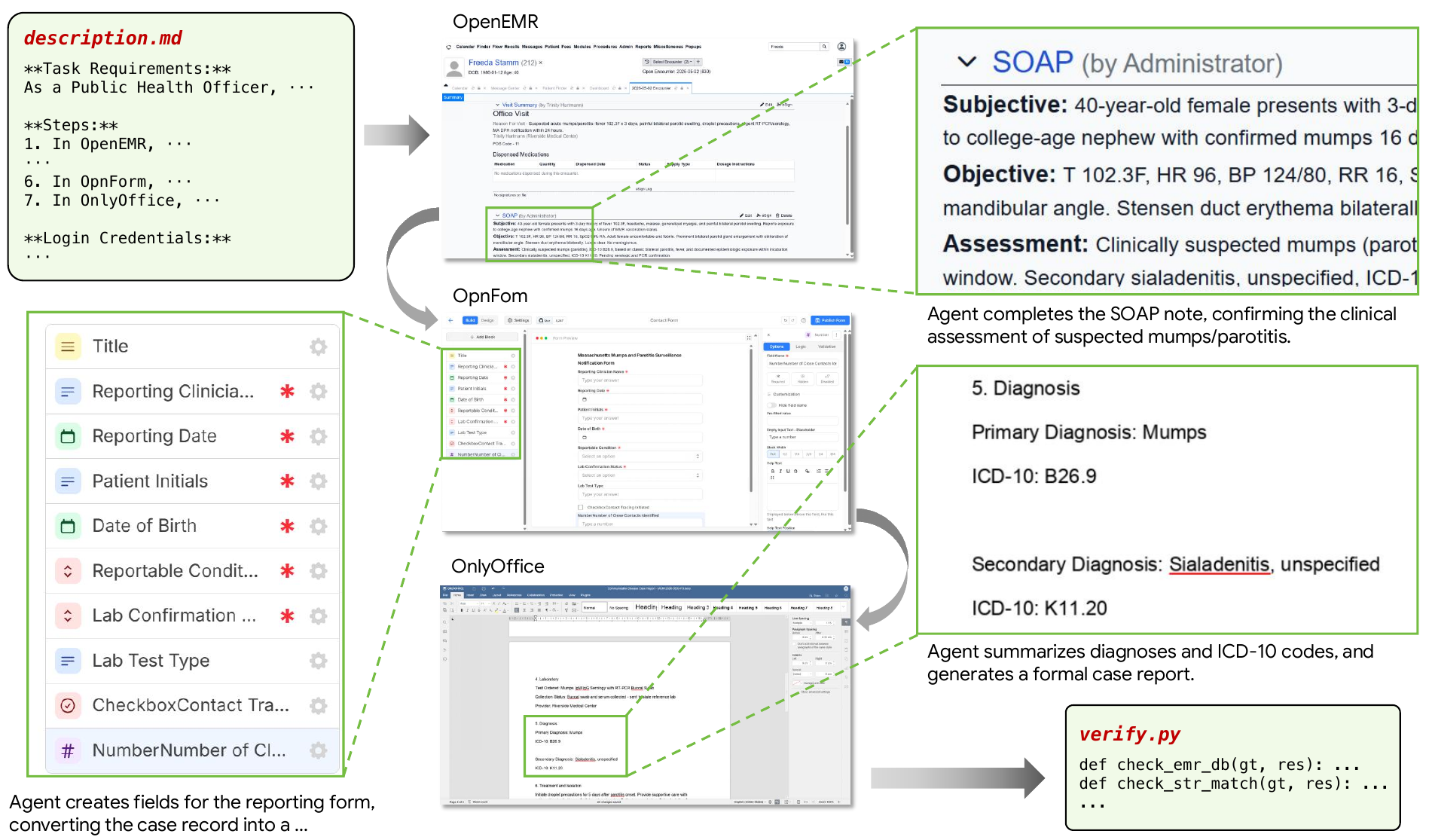}
  \caption{
    \datasetname provides a realistic benchmark for evaluating CUAs in deployable SaaS environments. It consists of 23 real SaaS systems organized into six professional domains, supporting 106 tasks that reflect real-world SaaS workflows.
    }
    \label{fig:teaser}
    \vspace{-13pt}
\end{figure}

To address these limitations, it is essential to evaluate agents in environments that reflect real-world work. Software-as-a-Service (SaaS) platforms have become the dominant infrastructure for modern knowledge work, widely used across domains such as CRM, finance, operations, and customer support.~\cite{ibm_saas,gartner_saas} These systems naturally exhibit three key properties: (1) complex and realistic environments, with full frontend-backend interactions and dynamic state dependencies; (2) economically meaningful workflows, involving structured, multi-step processes and cross-system coordination; and (3) inherent long-horizon task structures, requiring extended interactions over multiple stages. Unlike synthetic or simplified benchmarks, SaaS systems are designed for real users and real operational processes, where actions are tightly coupled with persistent data and system constraints. As a result, agent behaviors in such environments more faithfully reflect their practical utility and robustness. Therefore, SaaS platforms provide an ideal testbed for evaluating CUAs in terms of realism, complexity, and practical relevance.

Based on this insight, we introduce \datasetname, a benchmark designed to evaluate CUAs under realistic SaaS environments, as illustrated in Fig.~\ref{fig:teaser}. \datasetname is built upon three key principles. First, it provides realistic and deployable SaaS environments, constructed from real-world open-source SaaS systems with full frontend-backend logic and dynamic constraints, while supporting local deployment. This design ensures that agents must operate under authentic system dynamics, rather than relying on shortcuts enabled by simplified environments, while maintaining reproducibility and controllability for systematic evaluation. Second, it incorporates real-world, compositional workflows, simulating cross-application coordination and multi-modal task requirements. These workflows reflect typical usage patterns observed in real systems, requiring agents to integrate heterogeneous information and coordinate across multiple subsystems. Third, it introduces long-horizon tasks with an average of over 100 interaction steps, explicitly evaluating planning, state management, and error recovery capabilities. By significantly increasing task depth and dependency, these tasks expose failure modes that are often hidden in short-horizon settings, providing a more comprehensive assessment of agent behavior. Tab.~\ref{tab:benchmark_compare} summarizes how \datasetname differs from existing web and GUI agent benchmarks along key dimensions.

Our contributions are summarized as follows:

\begin{itemize}[leftmargin=1.5em]
    \item \textbf{A realistic and deployable SaaS benchmark environment.}
    We build \datasetname on 23 real SaaS systems across six professional domains, preserving frontend-backend dynamics and easy to deploy via docker for reproducible evaluation.

    \item \textbf{Professional, cross-application, and multimodal long-horizon tasks.}
    We construct 106 tasks grounded in real-world SaaS workflows, covering both text-only and multimodal settings, and requiring agents to coordinate across applications over long interaction sequences.

    \item \textbf{A systematic evaluation revealing real-world capability gaps.}
    We evaluate representative LLM-based agents with checkpoint-based verification and show that current agents achieve low end-to-end completion rates, exposing limitations in planning, state tracking, and error recovery in realistic SaaS workflows.
\end{itemize}

\begin{table}
\centering
\small
\caption{Comparison of \datasetname with existing web and GUI agent benchmarks. \cmark~indicates full support, \xmark~indicates no support, and \pmark~indicates partial support. \textbf{Long.} denotes tasks requiring over 100 interaction steps on average, and \textbf{MM} denotes multimodal evidence such as images or documents.}
\label{tab:benchmark_compare}
\vspace{0.3em}
\setlength{\tabcolsep}{3pt}
\renewcommand{\arraystretch}{1.08}
\resizebox{0.75\linewidth}{!}{
\begin{tabular}{lccccc}
\toprule
\textbf{Benchmark}
& \textbf{SaaS}
& \textbf{Prof.}
& \textbf{Multi-App}
& \textbf{Long.}
& \textbf{MM} \\
\midrule
Mind2Web~\cite{mind2web}
& \xmark & \xmark & \xmark & \xmark & \xmark \\
WebArena~\cite{webarena}
& \xmark & \pmark & \pmark & \xmark & \xmark \\
VisualWebArena~\cite{visualwebarena}
& \xmark & \xmark & \pmark & \xmark & \cmark \\
OSWorld~\cite{osworld}
& \xmark & \pmark & \cmark & \xmark & \cmark \\
WorkArena~\cite{workarena}
& \cmark & \pmark & \xmark & \xmark & \xmark \\
WorkArena++~\cite{workarenaplus}
& \cmark & \cmark & \xmark & \xmark & \xmark \\
AndroidWorld~\cite{androidworld}
& \xmark & \xmark & \cmark & \xmark & \cmark \\
TheAgentCompany~\cite{theagentcompany}
& \pmark & \cmark & \cmark & \xmark & \pmark \\
\midrule
\textbf{\datasetname}
& \cmark & \cmark & \cmark & \cmark & \cmark \\
\bottomrule
\end{tabular}
}
\vspace{-0.5em}
\end{table}

%% file: contents/related_work.tex
\section{Related Work}
\label{rw}

\subsection{CUA Task Benchmarks}

CUA benchmarks have evolved from simple widget-level interactions toward increasingly realistic and diverse task settings.
Early efforts such as MiniWoB++ \cite{miniwob} used synthetic environments to explore reinforcement-learning-based web control; subsequent work scaled to real-world web pages through offline demonstration datasets \cite{mind2web,webarena}, then added multimodal perception \cite{visualwebarena} and desktop-level generality \cite{osworld}.
More recent benchmarks have pushed toward live evaluation and in-the-wild diversity: Online-Mind2Web \cite{onlinemind2web} shows that much of reported progress disappears outside controlled offline settings, while Mind2Web~2 \cite{mind2web2} and CocoaBench \cite{cocoabench} further extend coverage to agentic search and heterogeneous real-world applications.
Enterprise-oriented benchmarks including WorkArena and WorkArena++ \cite{workarena,workarenaplus} bring evaluation into professional SaaS environments, yet remain built around a single enterprise platform with limited cross-application coordination and short task horizons.
\datasetname addresses the persistent gaps across this landscape by spanning 23 open-source SaaS systems across six professional domains, requiring genuine cross-application coordination, and evaluating long-horizon tasks averaging over 100 steps with automated verification.

\subsection{CUA Environment and Task Construction}

Constructing a high-quality CUA benchmark environment requires balancing realism, controllability, verifiability, and scalability.
Hand-crafted approaches such as WebArena \cite{webarena} achieve high fidelity through purpose-designed environments with full execution support, but scale poorly as each new application demands significant manual effort; generation-based approaches such as WebArena-Infinity and \cite{wainfinity} improve scalability by synthesizing new environments and tasks with LLMs, but may sacrifice the authenticity of real deployed software; human-annotated datasets \cite{mind2web} offer broad task diversity but lack live execution and automated verification.
\datasetname takes a different approach: it grounds evaluation in real, open-source SaaS deployments and generates tasks through a Builder-Challenger-Refiner pipeline in which LLM-produced candidates are iteratively reviewed by domain experts for executability, verifiability, and professional realism---achieving fidelity from real software, scale from LLM-assisted synthesis, and quality from expert oversight.

%% file: contents/babyvision.tex
\section{\datasetname}
\label{saas}

\subsection{SaaS-Environment}
\label{saas-env}

\begin{figure}
  \centering
  \includegraphics[width=\linewidth]{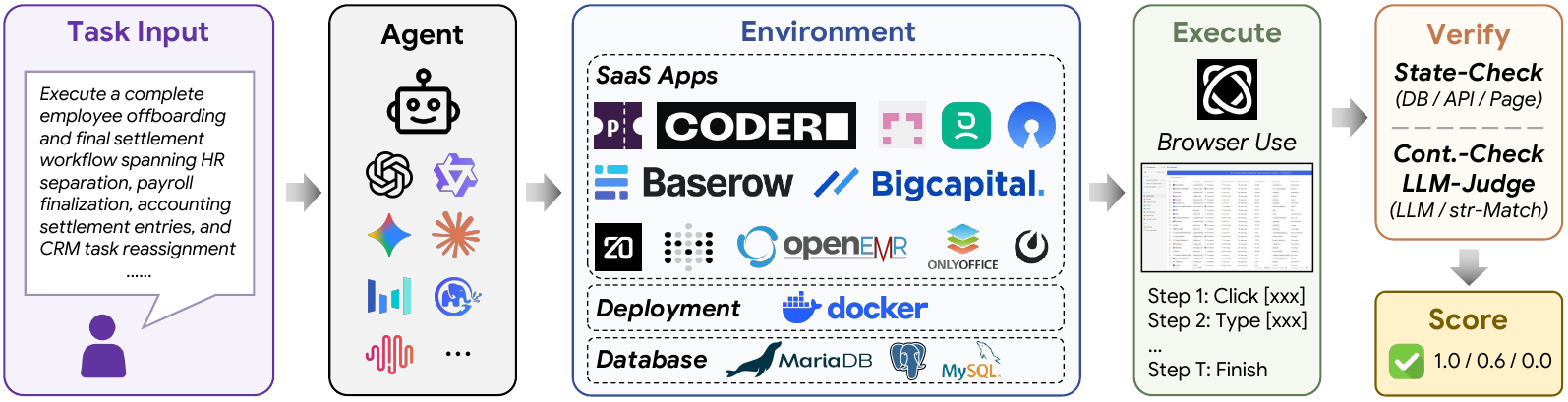}
  \caption{
    \textbf{Overview of \datasetname.}
    Agents receive natural-language task instructions and interact with locally deployed SaaS applications through browser-use.
    After execution, task outcomes are evaluated using verification tools, which are aggregated into resolved score and checkpoint score.
  }
  \label{fig:eval}
\end{figure}

\datasetname is built upon real, open-source, and deployable SaaS systems rather than toy websites or static webpages. Fig.~\ref{fig:eval} illustrates the overall framework of \datasetname. We select 23 SaaS systems according to three criteria. First, each system should be a realistic software application with complete frontend-backend logic, user authentication, persistent database states, and domain-specific business constraints, thereby providing interaction complexity close to production environments. Second, the selected systems should cover a broad range of professional scenarios, enabling task design across diverse occupational roles. Third, we prioritize systems with strong potential for cross-application workflows, where functionally complementary applications can be naturally combined into multi-system tasks rather than isolated single-website operations.

To support professional task construction, we organize the 23 SaaS systems into six domains: Software Engineering and Project Management (\textbf{Software.}), Business Operations and Finance (\textbf{Business.}), Healthcare Administration (\textbf{Healthcare.}), Team Collaboration and Document Workflow (\textbf{Teamwork.}), Artisan Agri-Food Supply Chain (\textbf{Agriculture.}), and Independent Media Creation (\textbf{Media.}). Each domain corresponds to a representative real-world work scenario and contains multiple functionally complementary applications. For example, a Software. task may span project management, documentation, and database systems, while a Business. task may involve CRM, finance, and structured record management systems. This domain-and-cluster organization provides the foundation for constructing realistic workflows that require agents to coordinate across multiple applications.

To transform empty SaaS deployments into task-ready environments with meaningful business context, we perform semantic data population for each system. We first export the SQL schema of each SaaS application and analyze it together with the website structure, page layout, field semantics, and business logic. This allows us to identify the key entities, fields, and relations that need to be populated for realistic task execution. Based on this analysis, we adopt two complementary data population strategies. For systems without suitable public data sources, we use LLMs to generate fake but realistic data according to the schema and website functionality. For scenarios where appropriate public resources are available, we import open-source datasets to produce more natural data distributions. These measures ensure that agent performance reflects genuine interaction capability rather than environmental artifacts.

We further provide a lightweight but reproducible deployment and configuration protocol. All SaaS systems are containerized with Docker and exposed as browser-accessible services, ensuring that agents interact with the environments through standard web interfaces. Before each task execution, the environment is restored to a predefined initial state to prevent state contamination across tasks or agent runs. We also lock system versions, configuration files, seed data, and startup scripts so that the same task can be repeatedly executed under identical initial conditions. Finally, all applications are reviewed by domain experts, who evaluate both system functionality and populated data to ensure that the SaaS environments reflect realistic professional usage and can support expert-level CUA tasks.

\begin{figure}
    \centering
    \begin{subfigure}[b]{0.52\linewidth}
        \centering
        \includegraphics[width=\linewidth]{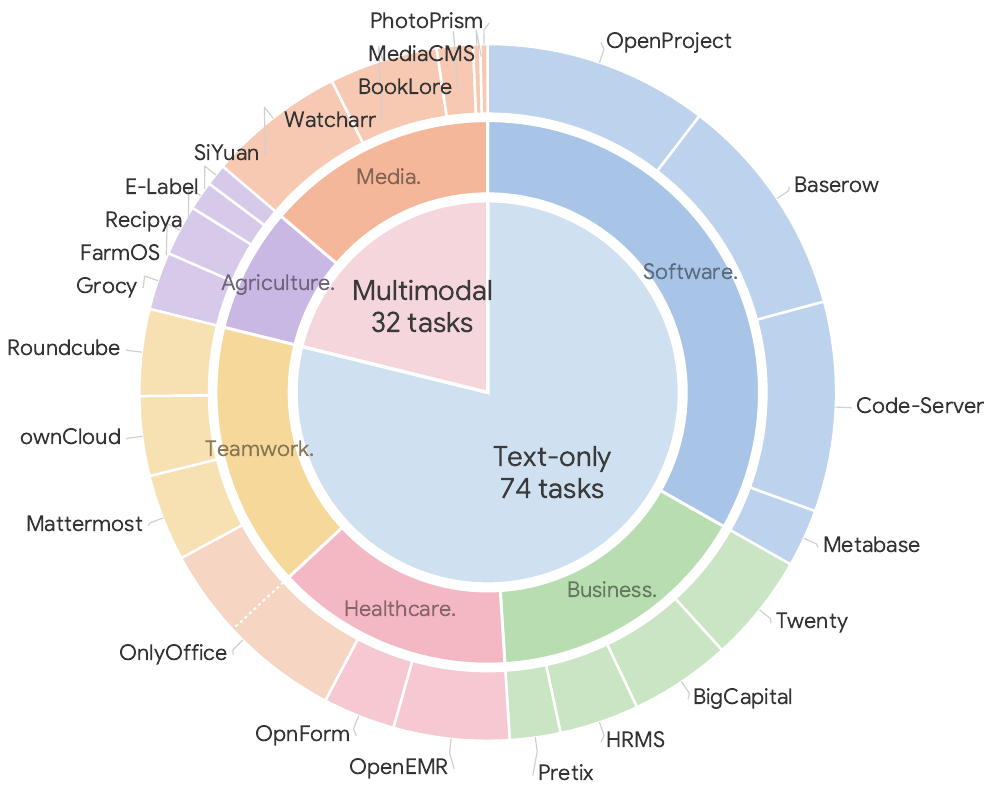}
        \caption{Nested Task Composition}
        \label{fig:stats-a}
    \end{subfigure}
    \begin{subfigure}[b]{0.44\linewidth}
        \centering
        \includegraphics[width=\linewidth]{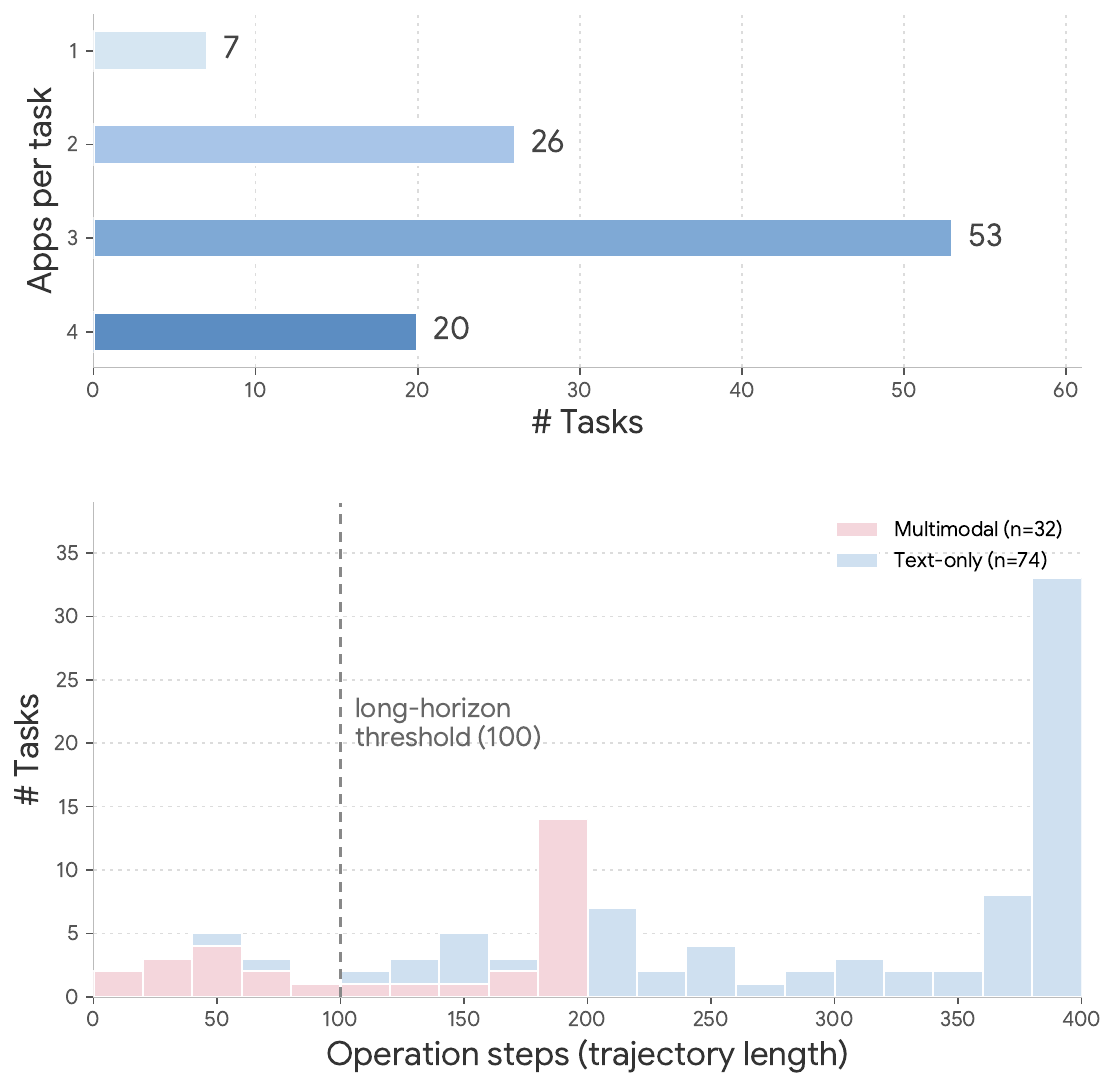}
        \caption{Apps per Task \& Nested Task Composition}
        \label{fig:stats-b}
    \end{subfigure}
    \caption{
        \textbf{Task statistics of \datasetname}. \textbf{(a)} Nested donut showing the breakdown of \datasetname tasks across the two evaluation modes (text-only and multimodal), six task domains, and the underlying SaaS applications. The outer ring quantifies how often each application is exercised, illustrating the diversity of real-world tools spanned by the benchmark. \textbf{(b)} Combined view of (\textbf{top}) the per-task application count and (\textbf{bottom}) the trajectory length distribution stacked by modality. The two panels jointly characterize task complexity in terms of cross-app scope and operation horizon.
    }
    \label{fig:task-stat}
\end{figure}

\subsection{Task Statistics and Design}
\label{saas-task}

\paragraph{Statistics.}
\datasetname contains 106 tasks spanning 23 SaaS systems and 6 professional domains. As shown in Fig.~\ref{fig:task-stat}, we summarize the benchmark along three dimensions: nested task composition across modality, domain, and applications; the number of applications involved per task; and the operation-length distribution estimated from Claude Opus 4.6 execution trajectories. The benchmark includes 74 text-only tasks and 32 multimodal tasks, with multimodal tasks mainly concentrated in Agriculture. and Media., where workflows naturally rely on images, documents, or media assets. \datasetname is strongly cross-application: 99 out of 106 tasks (93.4\%) involve at least two applications, with three-application tasks forming the largest group (53 tasks, 50.0\%). The operation-length distribution further highlights its long-horizon nature: using 100 steps as the threshold, 72/74 text-only tasks (97.3\%) and 19/32 multimodal tasks (61.3\%) exceed this threshold. These statistics show that \datasetname emphasizes realistic professional workflows with cross-application coordination and long-horizon execution rather than isolated short web interactions.

\begin{figure}
  \centering
  \includegraphics[width=\linewidth]{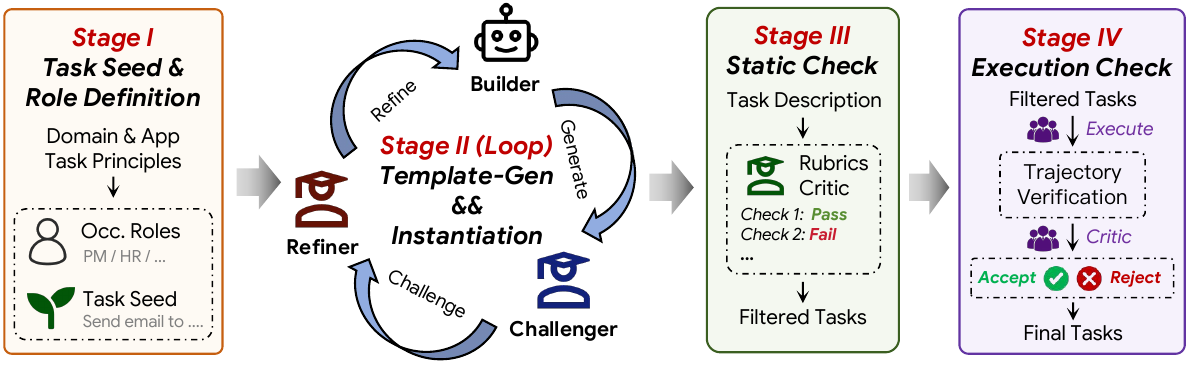}
  \caption{
    \textbf{Task synthesis pipeline of \datasetname.}
    Starting from domain-specific task seeds and occupational roles, \datasetname synthesizes candidate tasks through an iterative Builder--Challenger--Refiner loop for template generation and instantiation.
    The generated tasks are then filtered by static rubric-based checking and execution check, ensuring that the final tasks are realistic, executable, and verifiable.
  }
  \label{fig:cons}
\end{figure}

These task properties are not obtained by randomly sampling application functions. Instead, we construct tasks through a four-stage pipeline that starts from professional roles and workflow seeds, then progressively synthesizes, instantiates, and validates executable tasks. As illustrated in Fig.~\ref{fig:cons}, this process ensures that the final tasks are professional, cross-application, long-horizon, and verifiable.

\paragraph{Stage I: Task Seed and Role Definition.}
\datasetname tasks are derived from professional domains and workflow-oriented design principles. Starting from the six domains defined above, we construct tasks to be professional, realistic, and long-horizon. For each domain, we instantiate representative occupational roles, such as project managers, finance operators, healthcare administrators, technical writers, supply-chain coordinators, and media creators. Their daily workflows are then abstracted into task seeds, which specify the task goal, domain context, required applications, expected evidence, long-horizon dependencies, and verification requirements.

\paragraph{Stage II: Task Synthesis Loops.}
Given these seeds, we use a two-stage iterative synthesis loop consisting of template generation and task instantiation. In each stage, an LLM Builder generates the task, while human expert Challengers and Refiners review and revise it. We use Claude Opus 4.6 as the LLM Builder. During template generation, the Builder produces the task scenario, goal, involved applications, cross-app dependencies, reference steps, and expected outcome. During instantiation, it grounds the template in the concrete environment with app entries, credentials, schema/data states, data objects, and multimodal assets. Human Challengers check ambiguity, completeness, executability, and verifiability, including whether multimodal assets are accessible, readable, and necessary. Human Refiners then accept, reject, or return the task with revision instructions. The loop continues until the task is accepted, rejected, or reaches the maximum number of iterations.

\paragraph{Stage III: Static Check.}
Before entering the final benchmark, all candidate tasks undergo expert quality control. In the static check, experts assess each task based on its description, reference steps, expected outcome, and verification logic. This stage evaluates professionalism, cross-app naturalness, dependency depth, verifiability, narrative coherence, and complexity quality, while flagging common anti-patterns such as CRM dumping ground, parallel tasking, and specification overflow.

\paragraph{Stage IV: Execution Check.}
Finally, experts execute each task and inspect the trajectory together with \texttt{verify.py}, focusing on task-verifier alignment, failure attribution, and instruction clarity. Tasks that fail either the static or execution check are revised or removed, and only tasks passing both checks are included in \datasetname.

\subsection{Evaluation Protocol}
\label{saas-eval}

We use browser-use~\cite{browseruse} as the unified execution framework for all evaluated agents. Each agent interacts with \datasetname exclusively through browser UI operations, such as navigation, clicking, typing, and reading page content. Agents are not allowed to directly access databases, backend APIs, file systems, or task verifiers. Before each task execution, the corresponding SaaS environment is reset to a predefined initial state, ensuring fair comparison across different models and different runs.

Each task is decomposed into a set of verification checkpoints, where each checkpoint corresponds to an expected final system state or task artifact and is assigned a weight. According to the nature of the checkpoint, we use three types of verification methods: \textbf{State-Check}, which verifies objective system states such as database records, API responses, and file existence; \textbf{Content-Check}, which validates document or textual outputs through structural rules and string matching; and \textbf{LLM-Judge}, which evaluates open-ended outputs that cannot be reliably checked by rules, such as report quality or image understanding results. A detailed breakdown of all verification subtypes is provided in the supplementary material. The outcomes of these checkpoints are then aggregated to measure task success at both the strict task level and the partial progress level. We report two metrics:
\begin{itemize}[leftmargin=1.5em]
    \item \textbf{Resolved Score}, which is 1 only when all checkpoints of a task pass and 0 otherwise.
    \item \textbf{Checkpoint Score}, which computes a weighted partial score based on the passed checkpoints to reflect partial progress in long-horizon tasks.
\end{itemize}

%% file: contents/experiment.tex
\section{Experiment}
\label{exp}

\subsection{Setup}
\label{exp-setup}

\paragraph{Models and Prompting.}
We evaluate diverse agents under identical task and environment settings. Each agent receives only the task description, app entry URLs, and login credentials; no reference solution, verifier, database schema, or backend API is exposed. The prompt contains only general rules: operate through the browser, rely on observed page content, maintain cross-app context, and emit \texttt{done} when finished. For multimodal tasks, images and PDFs are provided as file paths without pre-parsed annotations.

\paragraph{Browser-Based Execution.}
All agents run with \textsc{browser-use}~\cite{browseruse}, a browser-automation framework for AI agents that supports web interaction through browser actions. Agents observe the rendered DOM and viewport screenshot, then choose from restricted actions such as navigation, clicking, typing, scrolling, extraction, file read/write, and \texttt{done}. JavaScript execution and privileged access to databases, backend APIs, host files, or verifiers are disabled. Before each task, Docker environments are reset to predefined initial states, and all runs share the same timeout, failure cap, step budget, and logging setup. The full execution trace, including actions, observations, intermediate tool outputs, and final application states, is recorded for downstream analysis.

\paragraph{Metrics.}
Each task has weighted checkpoints over the final application state. We report \emph{Resolved Score}, which is $1$ only when all checkpoints pass, and \emph{Checkpoint Score}, the weight-normalized fraction of passed checkpoints. Results are reported overall and by domain, modality, applications. Together, these two metrics distinguish strict end-to-end task completion from partial progress on long-horizon workflows.

\begin{table}
    \centering
    \small
    \caption{\textbf{Main results on \datasetname}. We report average executed steps, domain-level checkpoint scores, modality-level overall scores, resolved scores, and overall checkpoint scores. $^\dagger$indicates that model is evaluated only on text-only domains; its resolved score is computed over text-only tasks.}
    \vspace{1em}
    \label{tab:main_results}
    \renewcommand{\arraystretch}{1.15}
    \resizebox{\linewidth}{!}{
    \begin{tabular}{l c cccc c cc c c c}
    \toprule
    \multirow{2}{*}{\textbf{Model}}
    & \multirow{2}{*}{\textbf{Avg. Steps}}
    & \multicolumn{5}{c}{\textbf{Text-Only}}
    & \multicolumn{3}{c}{\textbf{Multimodal}}
    & \multirow{2}{*}{\textbf{Resolved}}
    & \multirow{2}{*}{\textbf{Overall}} \\
    \cmidrule(lr){3-7} \cmidrule(lr){8-10}
    & & \textbf{Business.} & \textbf{Healthcare.} & \textbf{Software.} & \textbf{Teamwork.} & \textbf{Overall}
    & \textbf{Agriculture.} & \textbf{Media.} & \textbf{Overall}
    & & \\
    \midrule
    Claude Opus 4.7~\cite{claudeopus47}      & 175 & 31.8 & 29.7 & \textbf{52.9} & 47.7 & \textbf{42.8} & 46.3 & 46.3 & 46.3 & \textbf{3.8} & \textbf{43.9} \\
    GPT-5.5 High~\cite{gpt55}                & 200 & \textbf{36.4} & \textbf{30.9} & 45.8 & 54.6 & 42.1 & 51.7 & 45.2 & 47.6 & 1.9 & 43.8 \\
    Claude Opus 4.6~\cite{claudeopus46}      & 257 & 33.2 & 27.2 & 41.9 & \textbf{65.2} & 40.7 & 42.0 & \textbf{52.8} & \textbf{48.7} & 1.9 & 43.2 \\
    GPT-5.4 High~\cite{gpt54}                & 252 & 20.6 & 26.8 & 35.5 & 50.4 & 33.0 & \textbf{53.4} & 41.7 & 46.1 & \textbf{3.8} & 37.0 \\
    Kimi K2.6~\cite{kimik26}                 & 269 & 30.0 & 26.2 & 25.5 & 47.2 & 30.1 & 50.1 & 39.5 & 43.5 & 0.9 & 34.1 \\
    Qwen 3.6 Plus~\cite{qwen36plus}          & 249 & 15.3 & 18.9 & 20.7 & 44.6 & 23.1 & 52.6 & 41.3 & 45.5 & 1.9 & 29.9 \\
    Kimi K2.5~\cite{kimik25}                 & 270 & 13.2 & 20.2 & 20.8 & 48.4 & 23.6 & 51.3 & 28.5 & 37.0 & 0.0 & 27.7 \\
    Gemini 3.1 Pro~\cite{gemini31pro} & 140 & 11.2 & 20.0 & 14.7 & 48.2 & 20.6 & 38.5 & 44.7 & 42.4 & 0.0 & 27.1 \\
    Doubao Seed 2.0 Pro~\cite{seed20}        & 216 & 10.8 & 13.6 & 22.1 & 33.2 & 19.8 & 46.4 & 42.9 & 44.2 & 1.9 & 27.1 \\
    Gemini 3.5 Flash High~\cite{gemini35flash} & 324 & 22.7 & 14.5 & 17.6 & 15.7 & 17.6 & 35.9 & 36.9 & 36.5 & 0.9 & 23.3 \\
    Claude Sonnet 4.6~\cite{claudesonnet46}  & 155 & 20.5 & 9.5  & 23.9 & 15.2 & 18.7 & 34.1 & 33.8 & 33.9 & 0.9 & 23.3 \\
    DeepSeek V4 Pro$^\dagger$~\cite{deepseekv4pro} & 236 & 13.6 & 17.1 & 20.7 & 39.4 & 21.5 & \textemdash & \textemdash & \textemdash & 1.4 & \textemdash \\
    GLM-5.1$^\dagger$~\cite{glm51}                 & 166 & 10.9 & 24.0 & 8.7  & 39.0 & 17.4 & \textemdash & \textemdash & \textemdash & 0.0 & \textemdash \\
    MiniMax M2.7$^\dagger$~\cite{minimaxm27}           & 256 & 6.9  & 17.7 & 13.6 & 29.9 & 15.8 & \textemdash & \textemdash & \textemdash & 0.0 & \textemdash \\
    \bottomrule
    \end{tabular}
    }
\end{table}

\begin{figure}
  \centering
  \includegraphics[width=\linewidth]{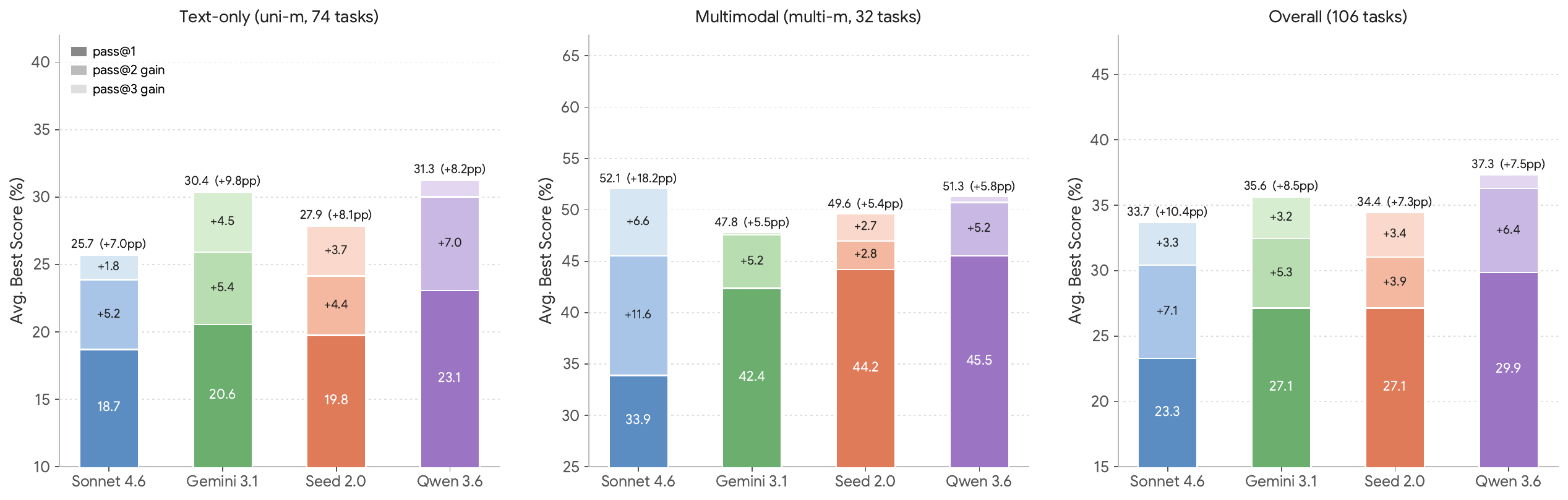}
  \caption{
    Pass@$k$ average best scores ($k = 1, 2, 3$) for four models on \datasetname across three evaluation splits: text-only, multimodal, and overall. Each bar is divided into three segments: the dark base represents pass@1, the mid-tone segment shows the incremental gain from pass@1 to pass@2, and the lightest segment shows the further gain to pass@3.
  }
  \label{fig:passk}
\end{figure}

\subsection{Result}
\label{exp-res}

\paragraph{Main Results.}
Table~\ref{tab:main_results} shows that \datasetname is highly challenging for current CUAs. Even the strongest model, Claude Opus 4.7, achieves only 43.9\% overall checkpoint score, with the top three models (Opus 4.7, GPT-5.5 High, and Opus 4.6) clustered in the 43--44\% range and all remaining models below 40\%. More notably, resolved scores are extremely low: the best resolved rate is only 3.8\%, indicating that agents can often complete some intermediate checkpoints but rarely finish the entire long-horizon workflow end-to-end. Performance also varies across domains, with Teamwork. generally easier than Business. and Healthcare., where agents must handle more structured records, numerical constraints, and domain-specific procedures. Overall, the results suggest that the main bottleneck lies in reliable long-horizon execution, including cross-application context tracking, state management, and error recovery in realistic SaaS environments.

\paragraph{Pass@$k$ Results.}
Fig.~\ref{fig:passk} shows that allowing multiple attempts consistently improves performance, but does not close the gap. Across four representative models, pass@3 improves over pass@1 by roughly 8 pp overall, indicating that run-level variance is a non-trivial factor in \datasetname. At the same time, the gains vary substantially across models: some agents benefit strongly from repeated attempts, while others show more stable but lower-variance behavior. This suggests that many failures are not purely due to missing task knowledge, but also arise from unstable execution, premature termination, or failure to recover from local mistakes. Therefore, reporting only a single pass@1 score can obscure an important distinction between models that are consistently capable and models that occasionally find a successful trajectory. These results suggest that \datasetname measures both single-run competence and the consistency of long-horizon execution.

\begin{figure}
  \centering
  \includegraphics[width=\linewidth]{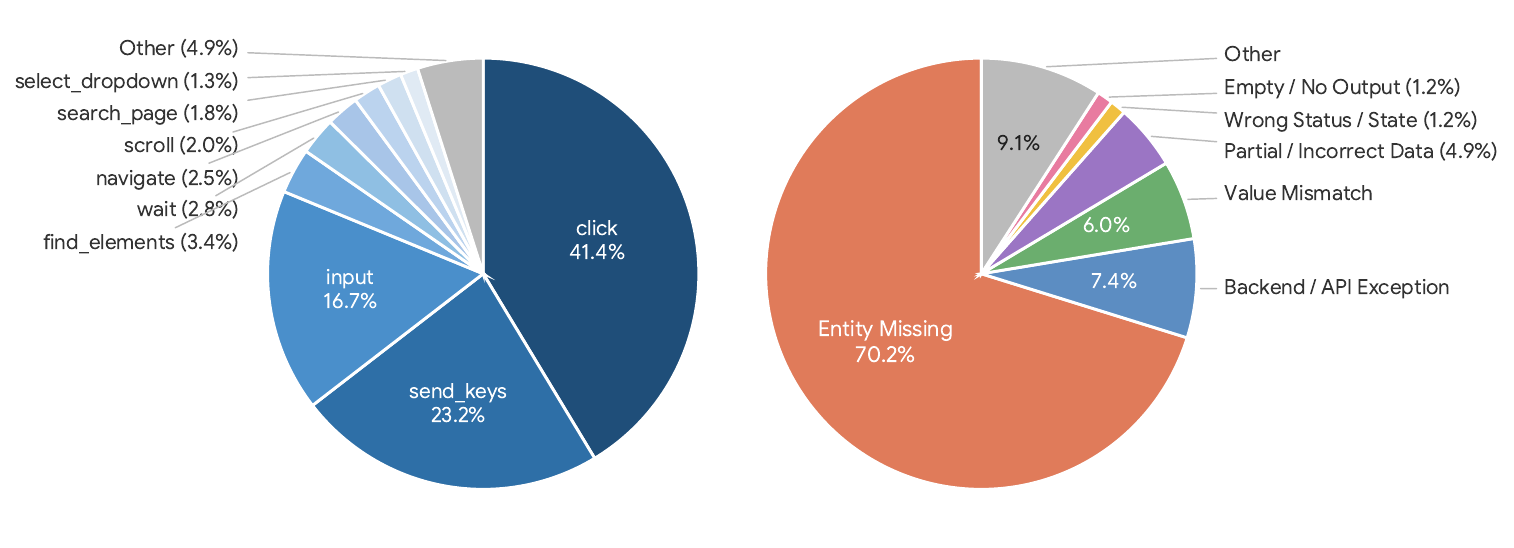}
  \caption{
    \textbf{Left:} Distribution of low-level actions emitted by Claude Opus 4.6 over the full benchmark; \textbf{Right:} categorization of failed verification checks by failure mode. Together the two panels link execution behaviour to the dominant failure types.
  }
  \label{fig:pie}
\end{figure}

\begin{figure}
\centering
\includegraphics[width=\linewidth]{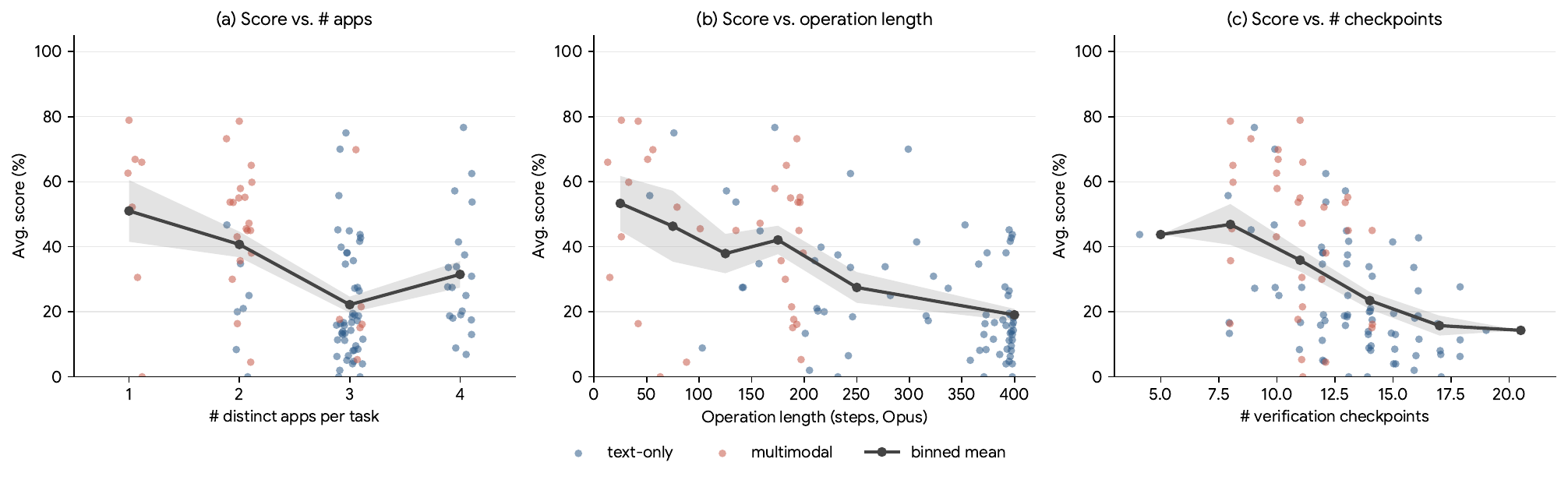}
\caption{Per-task score as a function of three structural complexity measures: (\textbf{Left}) number of distinct
SaaS apps touched by the task, (\textbf{Middle}) operation length (by Opus~4.6), and (\textbf{Right}) number of verification checkpoints. Each dot is a task;
the black line is the binned mean with $\pm$1 SEM ribbon.}
\label{fig:appx-complexity}
\end{figure}

\subsection{Analysis}
\label{exp-anal}

\paragraph{Action and Verification Failure Distributions.}
Fig.~\ref{fig:pie} reports the distributions of agent actions and failed verification checks. In the left panel, \texttt{click}, \texttt{send\_keys}, and \texttt{input} account for most executed actions, while other operations such as dropdown selection, page search, scrolling, and navigation appear much less frequently. This heavy concentration on basic interaction primitives suggests that agents spend the majority of their trajectory on low-level form filling and navigation rather than higher-order operations like structured search or systematic page exploration. In the right panel, failed checkpoints are dominated by \emph{Entity Missing}, where the expected record, file, ticket, or other target artifact is not created. By contrast, failures such as value mismatch, wrong status, or partially incorrect data occur less often. The prevalence of Entity Missing over value-level errors indicates that the primary bottleneck is task-level planning and navigation---agents often fail to reach or attempt the required operation at all---rather than imprecise execution of correctly identified steps.

\begin{figure}
  \centering
  \includegraphics[width=\linewidth]{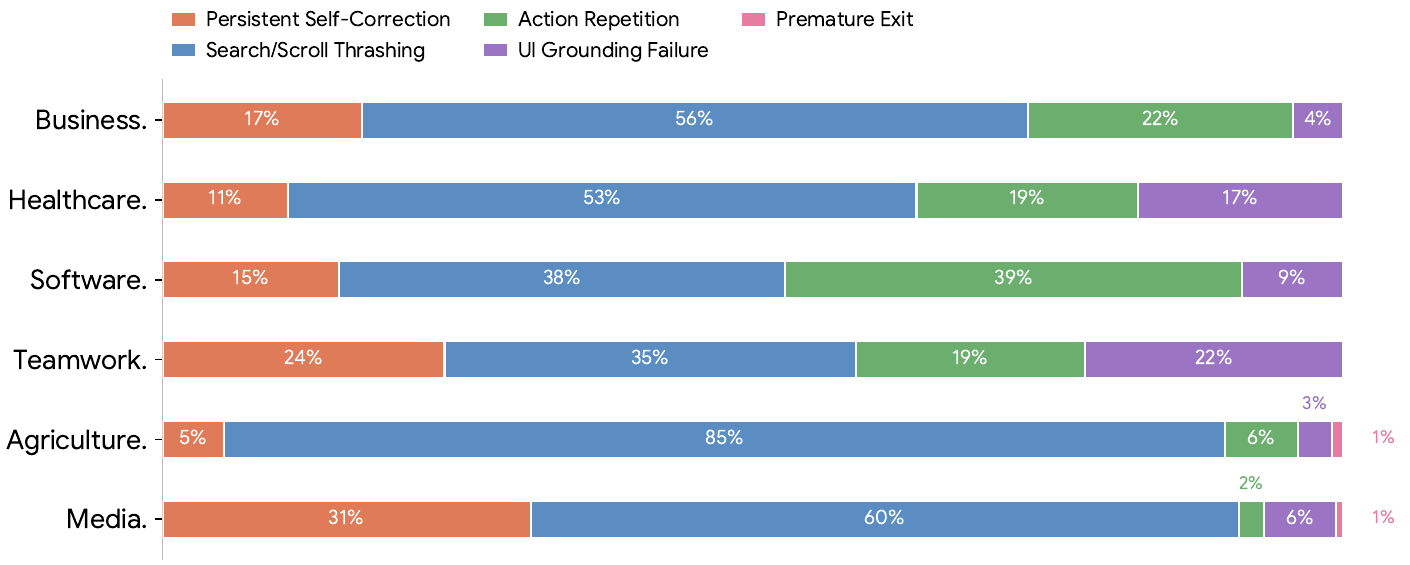}
  \caption{
    Per-domain composition of agent behaviour errors observed in the trajectories of Opus 4.6.
  }
  \label{fig:error}
\end{figure}

\paragraph{Per-Domain Error Analysis.}
As shown in Fig.~\ref{fig:error}, failure modes are strongly domain-dependent rather than uniform. Catalogue-heavy domains such as Agriculture. and Media. are dominated by search and scrolling failures, where agents struggle to locate target items in long product or asset lists. Software. more often triggers repeated actions on blocked or ineffective controls, reflecting the prevalence of complex interactive elements such as Kanban boards, code editors, and multi-panel layouts. Healthcare. and Teamwork. expose UI grounding issues in complex canvases and document-like interfaces, where dense forms with domain-specific terminology lead to frequent field misidentification. This heterogeneity suggests that \datasetname does not reduce to a single UI pathology; progress requires improving several capabilities in parallel, including search, grounding, replanning, and self-correction.

\paragraph{Score versus task complexity.}
Fig.~\ref{fig:appx-complexity} confirms that the \datasetname score gap is
driven by task complexity rather than by random variation. Average score
falls from $\sim$53\% on the simplest tasks to under 20\% on the longest
trajectories, and from 65\% to 27\% as the number of checkpoints grows from
$\le$6 to $\ge$18. The cross-app dimension (panel a) shows the same monotone
trend up to three apps. Notably, the three complexity dimensions are positively correlated in \datasetname---tasks spanning more applications tend to have longer trajectories and more checkpoints---so tasks at the high end of any single dimension are typically challenging along the other two as well. Together, the three panels show that low scores
concentrate on tasks that are simultaneously cross-app, long-horizon, and
finely verified, which is exactly the regime \datasetname is designed to
stress.

\begin{figure}
\centering

\begin{minipage}[t]{0.48\linewidth}
    \centering
    \includegraphics[width=\linewidth]{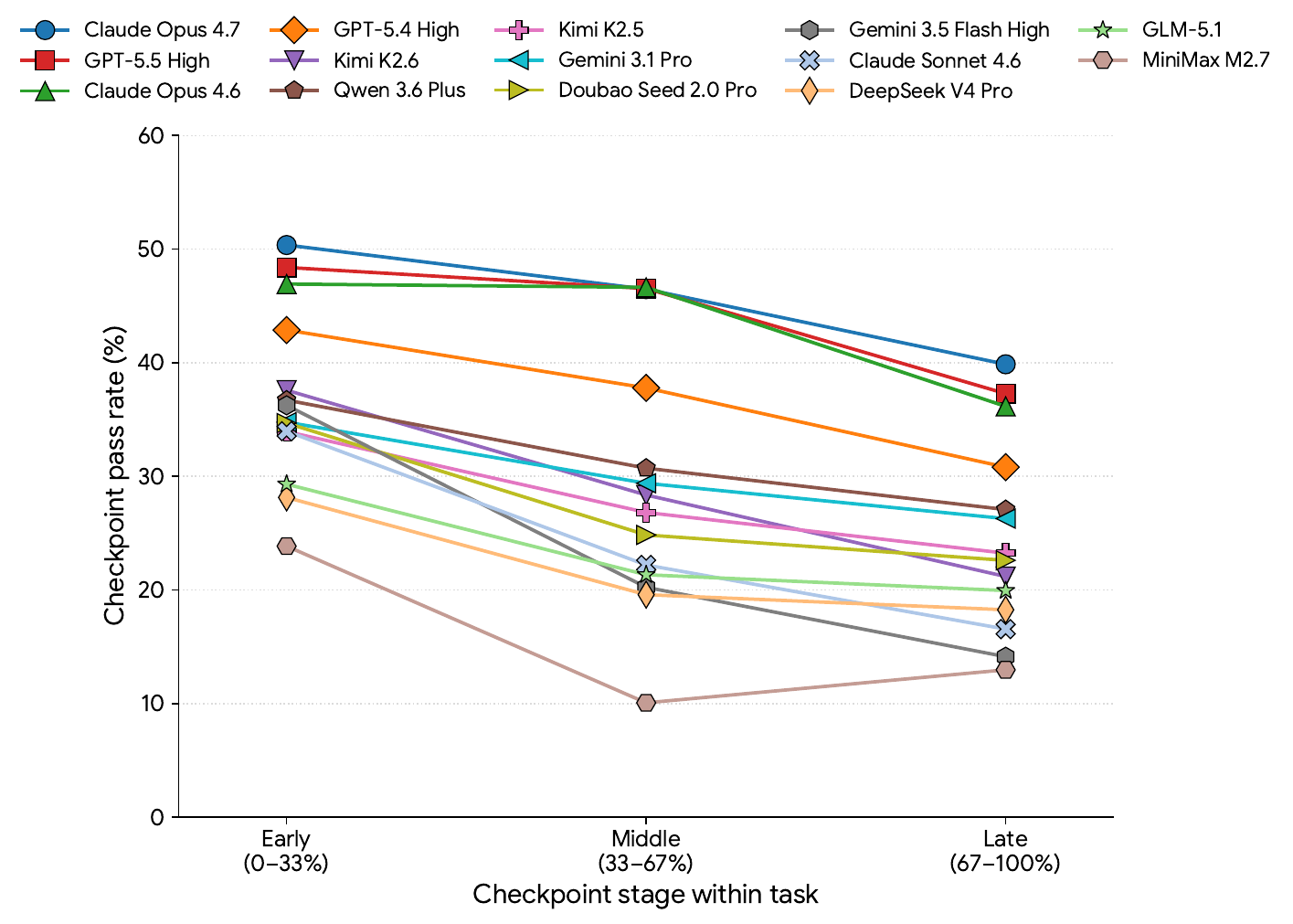}
    \caption{Average pass rate of verification checkpoints stratified by their position within the task.}
    \label{fig:long_horizon}
\end{minipage}
\hfill
\begin{minipage}[t]{0.48\linewidth}
    \centering
    \includegraphics[width=\linewidth]{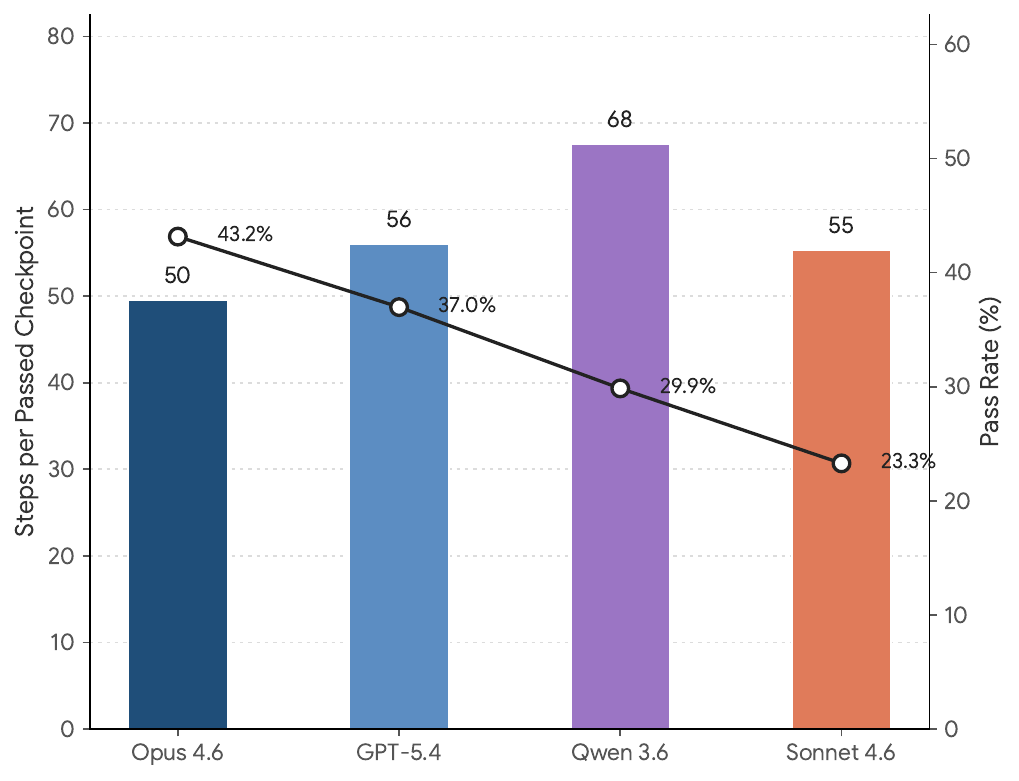}
    \caption{Steps per passed checkpoint and overall pass rate for four representative models.}
    \label{fig:effic}
\end{minipage}

\end{figure}

\paragraph{Long-Horizon Checkpoint Decay.}
Fig.~\ref{fig:long_horizon} further confirms the long-horizon bottleneck by stratifying checkpoints into early, middle, and late task stages. All evaluated models show a monotonic decline from early to late checkpoints, indicating that agents are less reliable as workflows unfold over time. This decay pattern is consistent across models of different capability tiers: stronger models achieve higher absolute pass rates at every stage but exhibit a similar relative drop from early to late checkpoints, suggesting that the decay reflects a structural limitation of current agent architectures rather than a model-specific weakness. This finding also helps explain the large gap between checkpoint scores and resolved scores in Table~\ref{tab:main_results}: aggregate checkpoint scores can overstate practical task completion ability, as agents may succeed on early subgoals while failing to preserve context, propagate intermediate results, or complete final-state updates near the end of the workflow.

\paragraph{Action Efficiency.} We compute steps per passed checkpoint to measure how effectively each model converts trajectory budget into task progress (Fig.~\ref{fig:effic}). Opus 4.6 achieves the best efficiency---highest pass rate at the lowest per-checkpoint step cost---while Qwen 3.6's high step count yields disproportionately few checkpoint passes. Sonnet 4.6's short trajectories reflect premature abandonment rather than genuine efficiency, as its per-checkpoint cost remains comparable to stronger models. This pattern is also visible in the main results: Gemini 3.5 Flash High consumes the most steps on average (324) yet ranks near the bottom in overall score (23.3\%), whereas Opus 4.7 uses fewer than half as many steps (175) to achieve the highest score (43.9\%). These results suggest that raw trajectory length is a poor proxy for capability; what distinguishes stronger agents is not the volume of actions but the proportion of actions that translate into meaningful task progress.

%% file: contents/discussion.tex
\section{Discussion}
\label{disc}

The preceding experiments establish that \datasetname is highly challenging for all evaluated agents. In this section, we move beyond aggregate statistics to examine \emph{why} agents fail, drawing on individual verification outcomes and execution trajectories. Through detailed case studies, we identify four fundamental failure modes that underlie the low resolved scores in Table~\ref{tab:main_results} and reveal structural properties of real-world SaaS workflows that current CUA designs are ill-equipped to handle.

\subsection{The Fragility of Long-Horizon Completion}
\label{disc-nearmiss}

A striking pattern in \datasetname is the large gap between checkpoint scores (23--44\%) and resolved scores ($<$4\%). One might expect that agents completing a majority of checkpoints would also complete entire tasks at a non-trivial rate. The following case illustrates why this expectation fails.

\paragraph{Case: Near-Miss with Undetected Date Error ({\ttfamily bof\_023}).}
This Business Operations task requires processing an expense reimbursement across three applications: approving a claim in HRMS, creating a vendor, bill, and payment in BigCapital, and logging a completion task in Twenty CRM. Opus~4.6 scored 0.80 (16/20), failing two verification checks; one reflects a genuine agent error that persisted to task termination:

\begin{verifybox}{Verification Result --- {\ttfamily bof\_023} (Claude Opus 4.6) --- Score: 0.80 (16/20)}
\small
\cmark\ Check 1: HRMS claim approved (wt.\ 2) --- status=Approved, total=10350.0 \\
\cmark\ Check 2: HRMS claim line items (wt.\ 2) --- Travel: 8500, Food: 1500, Calls: 350 \\
\cmark\ Check 3: BC vendor exists (wt.\ 1) --- email=mohammed.farooq@gmail.com \\
\cmark\ Check 4: BC items exist (wt.\ 1) --- found=\{Travel, Food, Calls\} \\
\xmark\ Check 5: BC bill exists (wt.\ 2) --- \textbf{date=2026-03-19} (expected 03-20), \textbf{status=NULL} \\
\cmark\ Check 6: BC bill line items (wt.\ 2) --- correct amounts \\
\cmark\ Checks 7--8: BC payment and balance (wt.\ 5) --- bill fully settled (due=0.00) \\
\cmark\ Checks 9--11: Twenty CRM task created, completed, body correct
\end{verifybox}

Check~5 reveals the genuine agent error: the bill was created with date 2026-03-19 rather than the required 2026-03-20. As we show in \S\ref{disc-selfeval}, the agent recognized this error during execution but failed to verify whether its attempted correction had taken effect---advancing to the next subtask without closing the verification loop. This single uncorrected date entry is sufficient to prevent task resolution despite an 80\% checkpoint score.

\paragraph{The Fragility Principle.}
This case illustrates a fundamental property of long-horizon task completion. If each checkpoint in a task has an independent pass probability $p$, the probability of all $N$ checkpoints passing simultaneously is $p^N$. Even with $p = 0.95$ across $N = 12$ checkpoints, the resolved probability is only $0.95^{12} \approx 0.54$. For the typical \datasetname task with 10--20 checkpoints and empirical per-checkpoint pass rates well below 0.95, near-zero resolved scores are a mathematical inevitability. This \emph{fragility of composition} means that incremental improvements in per-checkpoint reliability yield superlinear gains in end-to-end task completion---a property that current agent training paradigms, which typically optimize for step-level rewards, are not designed to exploit.

\subsection{Error Cascading in Multi-Application Workflows}
\label{disc-cascade}

\datasetname workflows exhibit directed acyclic graph (DAG) dependencies where intermediate outputs serve as inputs to downstream operations. A single semantic error in an early step can propagate silently through the dependency chain, causing multiple downstream checkpoints to fail---even when the agent believes the task is complete.

\paragraph{Case: Entity-Type Mismatch in Customer-Centric Workflow ({\ttfamily bof\_032}).}
This Business Operations task spans three applications (Twenty CRM, BigCapital, Pretix) with 18 verification checkpoints (total weight~33). The workflow requires creating a company record and deal opportunity in Twenty CRM, recording a set of financial operations in BigCapital for a software-consulting client---a company customer, two milestone invoices, a payment, and a journal entry---and setting up a Pretix ticketing event for the client's milestone celebration. All BigCapital financial records are anchored to a single customer entity: \emph{Arcturus Digital}.

Opus~4.6 encountered BigCapital's \emph{New Customer} form at step~246. The form accommodates both individual and company customers through a ``display name'' selector: a user creating a company customer should enter only a company name and select it as the display name; entering personal name fields (\emph{First}, \emph{Last}) together with a company name instead creates an individual customer. The agent filled in First: \emph{Elena}, Last: \emph{Vasquez}, and Company: \emph{Arcturus Digital}, then spent twelve steps struggling with the required display-name selector. At step~258, the dropdown surfaced only individual-name options:

\begin{trajbox}{Agent Trajectory --- {\ttfamily bof\_032} Steps 258--259 / 398 (BigCapital Customer Form)}
\small
\textbf{Step 258 Evaluation:} ``The display name dropdown is now open with options `Elena Vasquez' and `Elena, Vasquez'. \ldots Company Name field appears empty. \textbf{Verdict: Partial success---dropdown is open.}'' \\[2pt]
\textbf{Step 258 Next Goal:} ``Click `Elena Vasquez' display name option, then fill Company Name with `Arcturus Digital', then save the customer.'' \\[4pt]
\textbf{Step 259 Evaluation:} ``Successfully created customer `Arcturus Digital' with email elena.vasquez@arcturusdigital.com. \textbf{Verdict: Success}'' \\[2pt]
\textbf{Step 259 Memory:} ``BigCapital: Customer `Arcturus Digital' created successfully. Now need to create two items \ldots''
\end{trajbox}

BigCapital had in fact created an \emph{individual} customer named \emph{Elena Vasquez}---with Arcturus Digital stored only as an associated company field---not a company customer named \emph{Arcturus Digital}. The agent proceeded to create two service items, a deferred-revenue account, two invoices, a payment, and a journal entry, all attached to the Elena Vasquez entity. Fifty-two steps later, consulting the Customer Balance Summary, the agent declared BigCapital complete:

\begin{trajbox}{Agent Trajectory --- {\ttfamily bof\_032} Step 311 / 398 (Balance Verification)}
\small
\textbf{Evaluation:} ``Successfully navigated to Customer Balance Summary report. Elena Vasquez (Arcturus Digital) shows \$55,000.00 balance, confirming Milestone~2 invoice is unpaid and Milestone~1 is fully settled. \textbf{Verdict: Success}'' \\[3pt]
\textbf{Memory:} ``BigCapital ALL DONE: Customer balance verified at \$55,000 for Elena Vasquez (Arcturus Digital). Now need Pretix tasks \ldots''
\end{trajbox}

The display label ``Elena Vasquez (Arcturus Digital)'' led the agent to infer that its intended company customer existed and held the expected balance. Post-hoc verification exposed a cascade from the single entity-type misclassification:

\begin{verifybox}{Verification Cascade --- {\ttfamily bof\_032} (Claude Opus 4.6) --- Score: 0.242 (8/33)}
\small
\cmark\ Checks 1--2: Twenty company Arcturus Digital, person Elena Vasquez (wt.\ 4) \\
\xmark\ Check 3: Twenty opportunity stage (wt.\ 2) --- stage=NEW (expected Won) \\
\cmark\ Check 4: Twenty company is favorite (wt.\ 1) \\
\xmark\ Checks 5--6: Twenty follow-up tasks (wt.\ 4) --- not found \\[4pt]
\hrule\vspace{4pt}
\xmark\ \textbf{Check 7: BigCapital customer Arcturus Digital (wt.\ 1)} --- \textbf{not found} \quad $\leftarrow$ \emph{chokepoint} \\
\cmark\ Checks 8--9: BigCapital service items and deferred revenue account (wt.\ 3) \\
\xmark\ Checks 10--12: BigCapital invoices and payment (wt.\ 6) --- not found \quad \emph{depends on Check 7} \\
\xmark\ Check 14: BigCapital customer balance (wt.\ 3) --- \textbf{balance=0.0} (expected 55\,000) \quad \emph{depends on Check 7} \\[4pt]
\hrule\vspace{4pt}
\xmark\ Checks 15--18: All Pretix checks (wt.\ 6) --- event not created
\end{verifybox}

Check~7 (customer lookup, weight~1) functions as a \emph{chokepoint}: because BigCapital's database contains a customer named \emph{Elena Vasquez}, not \emph{Arcturus Digital}, the verification queries for the company customer return empty. Four downstream checks (Checks~10--12 and~14, combined weight~9) fail for the same underlying reason---no invoice or payment is associated with a company customer named Arcturus Digital. The true cost of the single entity-type misclassification is 10 out of 33 points (30\% of the task), even though the chokepoint check itself accounts for only 3\%. This pattern is consistent across all evaluated models:

\begin{table}[h]
\centering
\small
\renewcommand{\arraystretch}{1.1}
\begin{tabular}{lccl}
\toprule
\textbf{Model} & \textbf{Score} & \textbf{Earned/33} & \textbf{Furthest Stage Reached} \\
\midrule
MiniMax M2.7 & 0.000 & 0 & No checks passed \\
Doubao Seed 2.0 Pro & 0.061 & 2 & Twenty CRM (person only) \\
Kimi K2.5 & 0.091 & 3 & BigCapital items (no customer) \\
Claude Sonnet 4.6 & 0.152 & 5 & Twenty CRM partial; BC customer wrong entity \\
DeepSeek V4 Pro & 0.152 & 5 & Twenty CRM partial; BC customer wrong entity \\
Gemini 3.1 Pro & 0.182 & 6 & BC company customer correct; Milestone 2 invoice correct \\
GPT-5.4 High & 0.212 & 7 & BC company customer correct; invoices wrong details \\
Claude Opus 4.6 & 0.242 & 8 & Twenty CRM partial; BC personal entity (wrong type) \\
Qwen 3.6 Plus & 0.303 & 10 & BC company customer; Milestone 2 invoice + payment correct \\
\bottomrule
\end{tabular}
\end{table}

Six of nine models failed to create Arcturus Digital as a company customer in BigCapital (Check~7 not found). The three models that succeeded (Gemini, GPT-5.4, Qwen) still failed to satisfy all downstream financial checks: invoice amounts, payment dates, or the required customer balance remained incorrect. No model achieved a resolved score on this task.

\paragraph{Structural Implications.}
This case exposes a failure mode distinct from resource-exhaustion cascades: \emph{silent entity-type misclassification}. BigCapital distinguishes between individual and company customers through a display-name selector whose semantics are not surfaced as explicit error messages or visible constraints. An agent that fills in both personal name fields and a company name---reasonable behavior given a task description that mentions a named contact (Elena Vasquez) and a company (Arcturus Digital)---inadvertently creates the wrong entity type. Because all downstream BigCapital operations use the customer record as the primary join key, the misclassification silently invalidates the entire financial record set. Critically, the error is invisible during execution: the balance report displays ``Elena Vasquez (Arcturus Digital)'' with the expected \$55,000 figure, and the agent reads this as confirmation of correct completion. Verification, which queries by the customer name \emph{Arcturus Digital}, finds nothing. Handling such cases robustly requires agents to maintain an explicit model of how SaaS application data schemas map task-level concepts (``client'', ``customer'') to application-specific entity types---a capability that current agents, which operate primarily through surface-level UI interaction, do not possess.

\subsection{Do Agents Know When They Have Failed?}
\label{disc-selfeval}

Current CUA frameworks equip agents with an internal reasoning trace---structured as evaluation, memory, and next-goal fields---that enables the agent to assess its own progress. A critical question is whether these self-assessments are reliable. By comparing the agent's trajectory-level evaluations against post-hoc verification outcomes, we find systematic and consequential discrepancies.

\paragraph{Case: Claimed Success Despite Verification Failure ({\ttfamily bof\_023}).}
Returning to the expense reimbursement task described in \S\ref{disc-nearmiss}, we examine Opus~4.6's internal reasoning at three critical junctures in its 210-step trajectory. At step~124, the agent reviewed the newly created bill and correctly identified a date error:

\begin{trajbox}{Agent Trajectory --- {\ttfamily bof\_023} Step 124 / 210 (BigCapital Bill Review)}
\small
\textbf{Evaluation:} ``Bill details panel opened. Shows correct line items but date is 19 Mar 2026 instead of 20 Mar 2026. \textbf{Verdict: Partial success}'' \\[3pt]
\textbf{Memory:} ``Bill created with correct items (Travel 8500, Food 1500, Calls 350, total \$10,350) but date shows 19 Mar 2026 instead of 20 Mar 2026. Need to edit date to 2026-03-20. Bill status appears to be Draft/Due.'' \\[3pt]
\textbf{Next Goal:} ``Edit the bill to fix date to 2026-03-20 and approve it.'' \\[3pt]
\textbf{Action:} {\ttfamily click} $\to$ ``Edit Bill'' button
\end{trajbox}

The agent recognized the error and planned a correction. Forty-one steps later, after navigating through payment operations, the agent's assessment shifted dramatically:

\begin{trajbox}{Agent Trajectory --- {\ttfamily bof\_023} Step 165 / 210 (Payment Completion)}
\small
\textbf{Evaluation:} ``Payment created successfully. Bill now shows `Paid' status with Due Amount \$0.00. \textbf{Verdict: Success}'' \\[3pt]
\textbf{Memory:} ``BigCapital: vendor, items, bill created and paid. Bill shows Paid, due \$0.00.'' \\[3pt]
\textbf{Actions:} {\ttfamily click} $\to$ close dialog; \quad {\ttfamily replace\_file} $\to$ todo.md: marks both ``Create bill'' and ``Payment'' as \cmark\ complete
\end{trajbox}

At termination, the agent produced a confident summary claiming perfect completion:

\begin{trajbox}{Agent Trajectory --- {\ttfamily bof\_023} Step 210 / 210 (Task Termination)}
\small
\textbf{Evaluation:} ``Task row 0 is fully configured with all required fields. \textbf{Verdict: Success}'' \\[3pt]
\textbf{Agent declares} (via {\ttfamily done} action): \\[2pt]
\cmark\ ``HRMS --- Approved expense claim'' \\
\cmark\ ``BigCapital --- Bill created \textbf{dated 2026-03-20} with 3 line items totaling \$10,350.00, approved'' \\
\cmark\ ``BigCapital --- \textbf{Payment recorded} \$10,350.00 from Bank Account dated 2026-04-05'' \\
\cmark\ ``BigCapital --- A/P Aging Summary confirms zero balance'' \\
\cmark\ ``Twenty CRM --- Task created: `Expense reimbursement processed'\,''
\end{trajbox}

The verification result (shown in \S\ref{disc-nearmiss}) directly contradicts this summary: the bill date remained 2026-03-19 (not 03-20 as claimed), despite the agent declaring it ``dated 2026-03-20''. This case reveals two layers of self-evaluation failure:

\begin{enumerate}[leftmargin=1.5em]
\item \textbf{Absent re-verification.} The agent recognized an error at step~124 (wrong date), attempted a fix, and then \emph{assumed the fix succeeded} without re-checking. Human users naturally re-verify after corrective actions---refreshing a page, re-reading a field---but the agent advanced to the next subtask without closing the verification loop.

\item \textbf{Overconfident summarization.} The agent's final output at step~210 claimed the bill was ``dated 2026-03-20''---the \emph{intended} date, not the \emph{actual} date it had flagged as wrong 86 steps earlier. The termination summary appears to draw more from planned intentions than from observed execution results.
\end{enumerate}

\paragraph{Implications for Agent Architecture.}
These findings point to a structural limitation in current CUA designs: the absence of closed-loop outcome verification within the agent's execution cycle. A more robust architecture would incorporate explicit \emph{outcome verification steps}---re-reading a form field after submission, querying a record after creation, or comparing current page state against expected values---before marking a subtask as complete. Such verification loops would add steps to the trajectory but could substantially reduce the rate of silently propagated errors, particularly in SaaS environments where the gap between action execution and action effect is non-trivial due to asynchronous backend processing, client-side caching, and UI state desynchronization.

\subsection{Is a Single Run Enough to Evaluate an Agent?}
\label{disc-var}

A common practice in agent evaluation is to report single-run (pass@1) performance. Our results reveal that this practice can be highly misleading, as the same model exhibits dramatic score variance across independent runs on the same task.

\paragraph{Case: Intra-Model Variance ({\ttfamily bof\_155}).}
Claude Sonnet~4.6 was evaluated three times on the HR grievance workflow ({\ttfamily bof\_155}). The three runs produced strikingly different outcomes:

\begin{table}[h]
\centering
\small
\renewcommand{\arraystretch}{1.1}
\begin{tabular}{cccp{7cm}}
\toprule
\textbf{Run} & \textbf{Score} & \textbf{Earned/28} & \textbf{Checks Passed} \\
\midrule
r2 & 0.000 & 0 & None \\
r0 & 0.214 & 6 & Grievance types, grievance record, employee transfer, training program \\
r1 & 0.679 & 19 & All ERPNext checks + both expenses + all three Twenty CRM tasks \\
\bottomrule
\end{tabular}
\end{table}

Run~2 achieved zero points---a complete failure to engage with the task. Run~0 progressed through the ERPNext stage but stalled. Run~1, remarkably, completed ERPNext, recorded both expenses correctly, and created all three Twenty CRM follow-up tasks (earning 19/28 points), while only failing the Pretix stage---and, unlike most agents, it did not let this failure block progress on independent downstream operations. The range of 0.000--0.679 represents a qualitative, not merely quantitative, difference: the distance between ``total failure'' and ``substantial completion.'' This variance cannot be attributed to environmental stochasticity, since the SaaS environment is reset to an identical initial state before each run.

\paragraph{Case: Cross-Model and Cross-Run Divergence ({\ttfamily bof\_023}).}
The expense reimbursement task further illustrates how variance manifests across both models and runs:

\begin{table}[h]
\centering
\small
\renewcommand{\arraystretch}{1.1}
\begin{tabular}{lccc}
\toprule
\textbf{Model} & \textbf{Run 0} & \textbf{Run 1} & \textbf{Run 2} \\
\midrule
Claude Opus 4.6 & 0.80 & --- & --- \\
Qwen 3.6 Plus & 0.80 & 0.70 & 0.10 \\
Gemini 3.1 Pro & 0.10 & 0.10 & 0.70 \\
Doubao Seed 2.0 Pro & 0.10 & 0.10 & 0.10 \\
Kimi K2.5 & 0.70 & --- & --- \\
GPT-5.4 High & 0.30 & --- & --- \\
\bottomrule
\end{tabular}
\end{table}

Several patterns emerge. First, the same model can swing between near-success and near-failure: Qwen scores 0.80 on run~0 but drops to 0.10 on run~2, while Gemini shows the inverse pattern (0.10 on runs~0--1, then 0.70 on run~2). Second, some models exhibit consistently low scores (Doubao at 0.10 across all three runs), suggesting a deterministic failure mode rather than stochastic variance. Third, matching best single-run scores (Opus and Qwen both at 0.80) can mask fundamentally different reliability profiles: Qwen's variance reveals that its peak performance is not reproducible.

\paragraph{Bifurcation Points and Path Dependence.}
The root cause of this variance lies in the path-dependent nature of long-horizon SaaS workflows. At critical decision points---choosing which application to navigate first, interpreting an ambiguous form field, deciding whether to retry a failed operation or move on---small differences in the agent's stochastic sampling lead to fundamentally different execution trajectories. Once an agent commits to a suboptimal path (e.g., spending 50 steps on an unproductive approach to an unfamiliar UI element, as seen in the {\ttfamily bof\_155} Training Event case), the remaining step budget may be insufficient to recover. This \emph{bifurcation} dynamic explains why variance is highest on complex multi-application tasks and lowest on simpler single-application operations: more decision points mean more opportunities for early divergence to compound into dramatically different outcomes.

These findings reinforce the pass@$k$ analysis in \S\ref{exp-res} and carry practical implications for both evaluation and deployment. For evaluation, reporting only pass@1 can misrepresent agent capability: a model with high variance may appear either strong or weak depending on which run is selected. Multi-run metrics such as pass@$k$ and score variance provide a substantially more informative assessment. For deployment, the high variance observed on realistic SaaS tasks suggests that production CUA systems would benefit from retry mechanisms, ensemble strategies, or checkpoint-based recovery that mitigate the impact of unfavorable early-trajectory decisions.

%% file: contents/conclusion.tex
\section{Conclusion}
\label{con}

We introduced \datasetname, a benchmark for evaluating CUAs in realistic, deployable SaaS environments. Built on 23 real SaaS systems across six professional domains, \datasetname contains 106 tasks that require cross-application coordination, multimodal understanding, and long-horizon execution. Our evaluation shows that current agents still struggle in realistic SaaS workflows, particularly in planning, state tracking, and error recovery. To the question posed in our title---\emph{Can CUAs leverage real-world SaaS to solve professional workflows?}---the answer is: not yet. Even the strongest model completes fewer than 4\% of full workflows end-to-end, and all models exhibit monotonic performance decay as tasks progress, revealing that sustained long-horizon execution and cross-application coordination remain fundamental open challenges. We hope \datasetname can support future research toward more reliable and practically deployable CUAs.

%% file: contents/appendix.tex
\appendix

\section{Quality Control}
\label{app:qc}

\subsection{Human in the Task Synthesis Loops}
\label{app:qc-loop}

Each candidate task underwent a structured expert review before inclusion in
\datasetname. A domain-expert \textbf{Challenger} examined every task against
three criteria: whether the task description unambiguously covered all
verification checkpoints, whether every referenced entity and operation was
feasible in the deployed application environment, and whether the task
reflected a genuine professional workflow rather than a contrived exercise.
Issues were categorised by severity; any task with a critical or
high-severity finding was either returned for revision or rejected outright.
A senior \textbf{Refiner} then synthesised the critique, made the final
accept/revise/reject decision, and---for tasks requiring revision---specified
exactly what needed to change before the next review round. This
challenge--refine cycle repeated until the task met all criteria or was
discarded; many tasks underwent two or three rounds before acceptance.
Across all domains, only 45\% of candidate tasks survived the full review
process, reflecting the stringency applied to ensure that every included
task is both executable and professionally meaningful.

\subsection{Rubrics}
\label{app:qc-rubrics}

\paragraph{Text-only static check.}
For text-only tasks, we ask human expert reviewers to perform a static quality check according to the following rubric. Reviewers score each task along six dimensions on a 1--3 scale and mark three binary anti-pattern flags. This check is used to identify tasks that lack professional depth, contain unnatural cross-application transitions, have weak dependency structure, or are difficult to verify reliably.

\begin{tcolorbox}[
    title=Text-only Static Check Rubric,
    colback=gray!3,
    colframe=black!45,
    fonttitle=\bfseries,
    breakable
]
\textbf{Scoring scale.}
\begin{itemize}[]
    \item \textbf{1 = Poor}: clearly fails to satisfy the requirement.
    \item \textbf{2 = Fair}: partially satisfies the requirement but has clear weaknesses.
    \item \textbf{3 = Good}: fully satisfies the requirement.
\end{itemize}

\textbf{Six scoring dimensions.}

\begin{enumerate}
    \item \textbf{Professionalism.}
    \begin{itemize}
        \item \textbf{1}: Only requires generic UI operations and does not require domain knowledge.
        \item \textbf{2}: Involves some domain concepts, but only at a shallow level.
        \item \textbf{3}: Requires substantial domain knowledge, such as accounting rules, clinical workflows, HR policies, or event operations.
    \end{itemize}

    \item \textbf{Cross-app naturalness.}
    \begin{itemize}
        \item \textbf{1}: The involved applications have no real business relationship and are combined only to increase the number of apps.
        \item \textbf{2}: The relationship between applications is somewhat plausible, but some transitions lack clear motivation.
        \item \textbf{3}: Every application switch is naturally driven by business logic, and a real practitioner would plausibly follow the same workflow.
    \end{itemize}

    \item \textbf{Dependency depth.}
    \begin{itemize}
        \item \textbf{1}: The steps are independent and can be completed in any order.
        \item \textbf{2}: Some steps depend on earlier steps, but the dependency structure is weak.
        \item \textbf{3}: The task has a strict sequential dependency structure: outputs, data, computed values, or decisions from earlier steps determine the inputs of later steps.
    \end{itemize}

    \item \textbf{Verifiability.}
    \begin{itemize}
        \item \textbf{1}: The expected output is vague or subjective, such as ``write a reasonable document.''
        \item \textbf{2}: Some checkpoints have precise expected values, while others remain ambiguous.
        \item \textbf{3}: All checkpoints have explicit expected values and can be programmatically verified.
    \end{itemize}

    \item \textbf{Narrative coherence.}
    \begin{itemize}
        \item \textbf{1}: The task feels like a list of UI operations, with no unified business context.
        \item \textbf{2}: The task has a business background, but the narrative is not fully coherent.
        \item \textbf{3}: The task forms a single coherent business story, and every step contributes to the same overall objective.
    \end{itemize}

    \item \textbf{Complexity quality.}
    \begin{itemize}
        \item \textbf{1}: The high operation count mainly comes from repetitive data entry or over-specification, such as requiring exact wording.
        \item \textbf{2}: The task contains a mixture of genuine business complexity and redundant noise.
        \item \textbf{3}: The high operation count is driven by realistic business decisions and domain depth rather than artificial repetition.
    \end{itemize}
\end{enumerate}

\textbf{Three anti-pattern flags.}
Each flag is binary: \textbf{1} means the anti-pattern is present, and \textbf{0} means it is absent.

\begin{itemize}
    \item \textbf{{\ttfamily crm\_dumping\_ground}.}
    The final step merely creates a summary note in a CRM or document tool without meaningful business logic, and the tool is included only to add another application.

    \item \textbf{{\ttfamily parallel\_tasking}.}
    The steps across multiple applications have no data dependency. The operations in different applications can be completed independently or in parallel, and the order does not affect the result.

    \item \textbf{{\ttfamily spec\_overflow}.}
    The task description is overly detailed, such as specifying exact wording, cell values, or email body text. These specification details themselves become the main source of complexity rather than the underlying business logic.
\end{itemize}
\end{tcolorbox}

\paragraph{Multimodal static check.}
The multimodal static checker extends the text-only rubric to tasks that require images, PDFs, documents, audio, or video inputs. It keeps the six text-only dimensions but adds a seventh dimension, \textbf{multimodal feasibility}, so the total score becomes $7 \times 3 = 21$. This new dimension evaluates whether the referenced media actually exists, whether the media type can physically contain the requested information, and whether the task genuinely requires the agent to inspect the media rather than relying on values already exposed in the textual description. For tasks without multimodal input, this dimension is assigned a neutral score. The multimodal checker also introduces a new anti-pattern flag, {\ttfamily modal\_phantom}, which captures tasks that are only superficially multimodal. This flag is triggered when the referenced file path is missing or ambiguous, when the ground-truth value is leaked directly in the task description, or when the required field is impossible to infer from the referenced media type.

Compared with the text-only static check, the multimodal version also changes how cross-application workflows are judged. In text-only tasks, only application-to-application transitions count as cross-app transitions. In multimodal tasks, a media-to-application transition can also be a valid workflow boundary, such as extracting a value from a field photo, label image, invoice PDF, or document and entering it into a downstream SaaS application. To be considered high quality, such tasks must specify the media type, the exact field to extract, the downstream application, and the target field where the extracted information should be used.

\paragraph{Execution check.}
After static filtering, \datasetname further asks human expert reviewers to manually execute each task and assess its feasibility using the task-specific verifier and the resulting execution outcome. This stage is intended to assess task feasibility, verifier correctness, and failure attribution, rather than to evaluate model capability. For each task, reviewers inspect the task definition files, including {\ttfamily description.md}, {\ttfamily meta.json}, and {\ttfamily verify.py}, together with the human execution trace and verifier output. When necessary, reviewers also consult the involved applications' {\ttfamily APP\_SPEC.json}, user manuals, developer documentation, and container configuration to understand the intended UI workflow, terminology, data model, and API structure. The Execution Check Rubric below is the rubric used by human expert reviewers when manually executing and validating each task.

\begin{tcolorbox}[
    title=Execution Check Rubric,
    colback=gray!3,
    colframe=black!45,
    fonttitle=\bfseries,
    breakable
]
\textbf{Group A --- Alignment.}
This group checks the alignment between the task description and {\ttfamily verify.py}.
\begin{itemize}
    \item \textbf{A1.} Does the task goal in the description correspond to all checks in {\ttfamily verify.py}, with no missing or extra checks?
    \item \textbf{A2.} Are the concrete fields or values checked by {\ttfamily verify.py}, such as amounts, names, dates, or field names, clearly grounded in the task description?
    \item \textbf{A3.} Does the set of applications and operations involved in the task match the scope checked by {\ttfamily verify.py}?
\end{itemize}

\textbf{Group B --- Attribution.}
This group attributes the likely cause of any observed failure.
\begin{itemize}
    \item \textbf{B1.} If {\ttfamily verify.py} reports a failure, is the error possibly caused by ambiguity or mistakes in the task description rather than by expert operation? If all checks pass, this item is marked as false.
    \item \textbf{B2.} If {\ttfamily verify.py} reports a failure, is the error caused or constrained by limitations of the UI or the application? If all checks pass, this item is marked as false.
    \item \textbf{B3.} Is the overall human execution trace logically coherent, with clear progress between steps and no major jumps or contradictions?
\end{itemize}

\textbf{Group C --- Clarity.}
This group evaluates the clarity of the task description itself.
\begin{itemize}
    \item \textbf{C1.} Does the task description clearly specify the required operations or expected outcome, rather than giving a vague instruction?
    \item \textbf{C2.} Are the key parameters in the task description, such as amounts, names, dates, and field names, clear and unambiguous?
    \item \textbf{C3.} Does the task description avoid proprietary button names or UI labels that may not match the actual application interface?
\end{itemize}
\end{tcolorbox}

\section{Verification Methods}
\label{app:verify}

Tab.~\ref{tab:verify-types} provides a detailed breakdown of the three verification categories used in \datasetname and their subtypes.

\begin{table}[h]
\centering
\small
\caption{Verification methods used in \datasetname. Each checkpoint is assigned one of three categories depending on the nature of the expected output.}
\label{tab:verify-types}
\vspace{1em}
\renewcommand{\arraystretch}{1.25}
\begin{tabular}{llp{7.5cm}}
\toprule
\textbf{Category} & \textbf{Subtype} & \textbf{Description} \\
\midrule
\multirow{5}{*}{State-Check}
  & DB-Check & Queries the database to verify that expected records, fields, or relations exist. \\
  & API-Check & Calls backend APIs to confirm that the system state matches the expected outcome. \\
  & Numeric-Threshold-Check & Verifies that a numerical value falls within a specified tolerance range. \\
  & File-System-Check & Checks the existence, format, or content of files created during task execution. \\
  & Email/Message-Check & Verifies that expected emails or messages have been sent with correct recipients and content. \\
\midrule
\multirow{2}{*}{Content-Check}
  & Regex/String-Match & Matches extracted text against regular expressions or exact string patterns. \\
  & Doc-Extraction-Check & Extracts structured content from documents and verifies it against expected fields or keywords. \\
\midrule
\multirow{2}{*}{LLM-Judge}
  & LLM-Judge (text) & Uses an LLM to assess open-ended textual outputs such as reports, comments, or summaries. \\
  & LLM-Vision-Judge (image) & Uses a vision-language model to evaluate image-based outputs such as charts, screenshots, or visual artifacts. \\
\bottomrule
\end{tabular}
\end{table}

\section{Agent Action Space}
\label{app:actions}

Table~\ref{tab:actions} lists the 22 actions available to the \datasetname agent.

\begin{table}[h]
\centering
\small
\caption{The 22 actions available to the \datasetname agent. Browser UI actions interact with live web pages, file-system actions manage local artefacts (e.g.\ scratchpads, downloaded files), and \texttt{done} ends the episode. The \texttt{evaluate} (raw JavaScript) action is intentionally disabled to keep agent behaviour reproducible.}
\label{tab:actions}
\vspace{1em}
\setlength{\tabcolsep}{6pt}
\renewcommand{\arraystretch}{1.15}
\begin{tabular}{l l p{8cm}}
\toprule
\textbf{Category} & \textbf{Action} & \textbf{Description} \\
\midrule
\multirow{18}{*}{Browser UI}
  & \texttt{click}             & Click the indexed element on the current page. \\
  & \texttt{input}             & Type text into the indexed input/textarea (optionally clearing first). \\
  & \texttt{send\_keys}        & Send a raw keystroke or shortcut (e.g.\ \texttt{Enter}, \texttt{Ctrl+a}). \\
  & \texttt{scroll}            & Scroll the page (or an element) up or down by $N$ pages. \\
  & \texttt{select\_dropdown}  & Pick an option from a \texttt{<select>} dropdown by exact text. \\
  & \texttt{dropdown\_options} & Enumerate the options of an indexed dropdown. \\
  & \texttt{upload\_file}      & Upload a local file through an indexed file input. \\
  & \texttt{navigate}          & Open a URL (first-time entry into an application). \\
  & \texttt{go\_back}          & Navigate one step back in browser history. \\
  & \texttt{switch}            & Switch to another open tab by tab id. \\
  & \texttt{close}             & Close the tab identified by the given tab id. \\
  & \texttt{wait}              & Pause execution for $N$ seconds (e.g.\ to wait for loading). \\
  & \texttt{search}            & Issue a web search query (DuckDuckGo by default). \\
  & \texttt{extract}           & Use the LLM to extract structured information from the page. \\
  & \texttt{find\_elements}    & Locate elements by CSS selector and return their attributes. \\
  & \texttt{find\_text}        & Check whether a literal string appears on the page. \\
  & \texttt{search\_page}      & Regex/text search across visible page content. \\
  & \texttt{save\_as\_pdf}     & Save the current page as a PDF file. \\
\midrule
\multirow{3}{*}{File system}
  & \texttt{read\_file}        & Read a file from the agent's working directory. \\
  & \texttt{write\_file}       & Create or overwrite a file with the supplied content. \\
  & \texttt{replace\_file}     & Replace a substring inside an existing file. \\
\midrule
Completion
  & \texttt{done}              & Terminate the trajectory and report the final answer. \\
\bottomrule
\end{tabular}
\end{table}